\documentclass{article}

\usepackage{arxiv}

\usepackage[utf8]{inputenc} 
\usepackage[T1]{fontenc}    
\usepackage{hyperref}       
\usepackage{url}            
\usepackage{booktabs}       
\usepackage{amsfonts}       
\usepackage{nicefrac}       
\usepackage{microtype}      
\usepackage{lipsum}
\usepackage{times}
\usepackage{epsfig}
\usepackage{graphicx}
\usepackage{amsmath}
\usepackage{amssymb}
\usepackage{rotating}
\usepackage{multirow}
\usepackage{subfigure}
\usepackage{bm}
\usepackage{hhline}
\usepackage[affil-it]{authblk}
\usepackage[norule,symbol,perpage]{footmisc}
\begin{document}
\title{Review of the Fingerprint Liveness Detection (LivDet) competition series: from 2009 to 2021}

\author[1]{Marco Micheletto}
\author[1]{Giulia Orrù}
\author[1]{Roberto Casula}
\author[2]{David Yambay}
\author[1]{Gian Luca Marcialis}
\author[2]{Stephanie C. Schuckers}
\affil[1]{%
    University of Cagliari, Italy \{marco.micheletto,giulia.orru,roberto.casula,marcialis\}@unica.it}
\affil[2]{%
    Clarkson University, USA  \{yambayda,sschucke\}@clarkson.edu}

\maketitle
\begin{abstract}
Fingerprint authentication systems are highly vulnerable to fingerprint artificial reproductions, called fingerprint presentation attacks. Detecting presentation attacks is not trivial because attackers refine their replication techniques from year to year.
The  International  Fingerprint  liveness  Detection  Competition  (LivDet), an  open  and  well-acknowledged meeting point of academies and private companies that deal with the problem of presentation attack detection, has the goal to assess the performance of fingerprint presentation attack detection (FPAD) algorithms by using standard experimental protocols and data sets. Each LivDet edition, held biannually since 2009, is characterized by a different set of challenges to which the competitors must provide solutions
The continuous increase of competitors and the noticeable decrease in error rates across competitions demonstrate a growing interest in the topic. This paper reviews the LivDet editions from 2009 to 2021 and points out their evolution over the years.
\end{abstract}
\section{Introduction}
\label{intro}
In recent years, fingerprint-based biometric systems have achieved a significant degree of accuracy. This biometry can reasonably be considered the most mature from both an academic and an industrial point of view. Nowadays, it is one of the most suitable solutions in every application that requires a high level of security.
The reason for this success is mainly due to its universality, duration, and individuality properties. 
Nevertheless, it has been repeatedly shown that artificial fingerprints can fool a fingerprint-based personal recognition system \cite{matsu,Marcel}. As a matter of fact, most standard scanners on the market cannot distinguish images of a real fingerprint from the so-called ``spoofs''.
For this purpose, a fingerprint presentation attack detector (FPAD) is designed to automatically decide during usage about the liveness of the fingerprint presented to the sensor. In these systems, additional information is usually exploited to check whether the fingerprint is authentic or not. Liveness detection systems are therefore included in two categories: (1) hardware-based, which add specific sensors to the scanner to detect specific traits that ensure vivacity (e.g. sweat, blood pressure, etc.); (2) software-based, which process the sample acquired by a standard sensor to verify the liveness, by extracting distinctive features from the image and not from the finger itself.

Among the several initiatives, the International Fingerprint Liveness Detection Competition (LivDet) is a biennial meeting aimed at making the point on the limits and perspectives of presentation attacks detectors (PADs). The competition occurred and evolved in seven editions between 2009 \cite{livdet2009} and 2021 \cite{livdet2021}, proposing new challenging falsification methods, kind of scanner, and spoofs' materials. 

This paper examines all the editions of the LivDet competition, analyzing the progress of the liveness algorithms in twelve years.

Section 2 describes the background of spoofing and liveness detection. Section 3 describes the methods used in the competition test sets. Section 4 examines the results in terms of system accuracy. Section 5 discusses the analysis of the fingerprint's image quality. Section 6 concludes the paper.

\section{Fingerprint Presentation Attack Detection}
\label{sec:FPAD}
Since 1998, the year in which the vulnerability of personal recognition systems to artificial fingerprint replicas was demonstrated \cite{Scanners}, the scientific community has continued to study and propose new methods to protect such systems with the so-called Presentation Attack Detectors (PADs).
In the same way, the ability of attackers continues to evolve over the years and the study of attack techniques is the basis of research on Presentation Attack detection. It is therefore important to highlight the main techniques for creating a fake fingerprint, which can be consensual or non-consensual.
The consensual approach is regarded as the worst-case scenario, as the user's collaboration permits the creation of a high-quality fake. This category includes the basic \textit{molding and casting} method \cite {matsu} and sophisticated 2D and 3D printing techniques \cite{7903634}. Three phases are involved in the \textit{molding and casting} process. In the initial stage, the volunteer presses a finger into a silicone material, leaving a negative impression of her or his biometric trait on a mold. After that, the mold is filled with a casting substance, such as latex, liquid ecoflex, or glue. The solidified material is released from the mold and represents an exact replica of the genuine fingerprint, allowing a presentation attack against a fingerprint recognition system.
The non-consensual method involves greater risks for the user as it is more realistic in terms of attacks on a biometric system. 

Unlike the casts created in the consensual method with the collaboration of a volunteer, the latent imprint can be detected by an impostor through various techniques \cite{4563117}. By applying magnetic powders on a smooth or non-porous surface, it is possible to highlight the impression so that it can be photographed or scanned \cite{SODHI2001172}. This will be subsequently processed and the negative will be created and printed on a sheet. Another way to create the fake is to engrave the negative on a printed circuit in order to drip the material.

To defend against these types of attacks, various software solutions for liveness detection have been implemented over the years \cite{Marasco:2014:SAS:2658850.2617756}. As with other biometrics, presentation attack detection techniques have also undergone an evolution passing from the analysis of ridges and valleys, to local hand-crafted methods based on morphology, color and texture analysis such as BSIF and LBP \cite{10.1007/978-3-319-12484-1_21,9312046,8552931}, and to more modern deep-learning techniques \cite{8630773,8553413}.

\section{The Fingerprint Liveness Detection Competition}
\label{sec:livdet}
The Fingerprint Liveness Detection Competition was established with the goal of developing a standard for evaluating PAD technologies.

Born in 2009 \cite{livdet2009} through the collaboration of the University of Cagliari and Clarkson University, the number of participants and algorithms has steadily increased over the editions. Using a competition rather than merely sharing datasets might ensure free, third-party testing. Because of a conflict of interest, Clarkson and Cagliari have never competed in LivDet. 

The Algorithms part of the LivDet competition provides a common experimental protocol and data sets and evaluates the performance of FPAD systems with different edition-by-edition challenges that simulate open issues of the state of the art. Each competitor receives a set of images (training set), trains their FPADs with them and submits the solution. The organizers test the algorithms on a set of new images (test set) and calculate the performances and the final ranking. 

The interest in the Algorithms part of the LivDet competition and the number of participants and algorithms grows from edition to edition as shown in Table \ref{table:competition}.

\begin{table}[]
\centering
\caption{Number of competitors and algorithms submitted to the seven editions of the Fingerprint Liveness Detection Competition - Algorithm part.}
\label{table:competition}
\resizebox{0.7\textwidth}{!}{%
\begin{tabular}{|l|c|c|c|c|c|c|c|}
\hline
\multicolumn{1}{|c|}{\textbf{LivDet Edition}} & \textbf{2009} & \textbf{2011} & \textbf{2013} & \textbf{2015} & \textbf{2017} & \textbf{2019} & \textbf{2021} \\ \hline
\textbf{\# Participants}                      & 4             & 3             & 9             & 10            & 12            & 9             & 13            \\ \hline
\textbf{\# Algorithms}                        & 4             & 3             & 11            & 12            & 17            & 12            & 23            \\ \hline
\end{tabular}}
\end{table}

The Systems part of the LivDet competition originally began with one competition, but in 2017 a second competition was added. For both competitions, each competitor submits a fingerprint sensor trained on their own FPAD data, with some FPAD recipes supplied to competitors. The organizers then recruit live subjects and create PAIs to test the system based on which competition to which the system is submitted. The first competition is a PAD only module for systems, which detects if the input is live or spoof. The second competition is the incorporation of presentation attack detection with matching to determine overall performance of a system.  For this portion, a live person enrolls and a spoof is created of the same fingerprint of the enrolled person.  A successful attack bypasses the PAD module and matches the enrolled fingerprint.

\section{Methods and Dataset}
\label{sec:methods}
\subsection{Algorithms part}

\begin{figure}
    \centering
    \includegraphics[width=\textwidth]{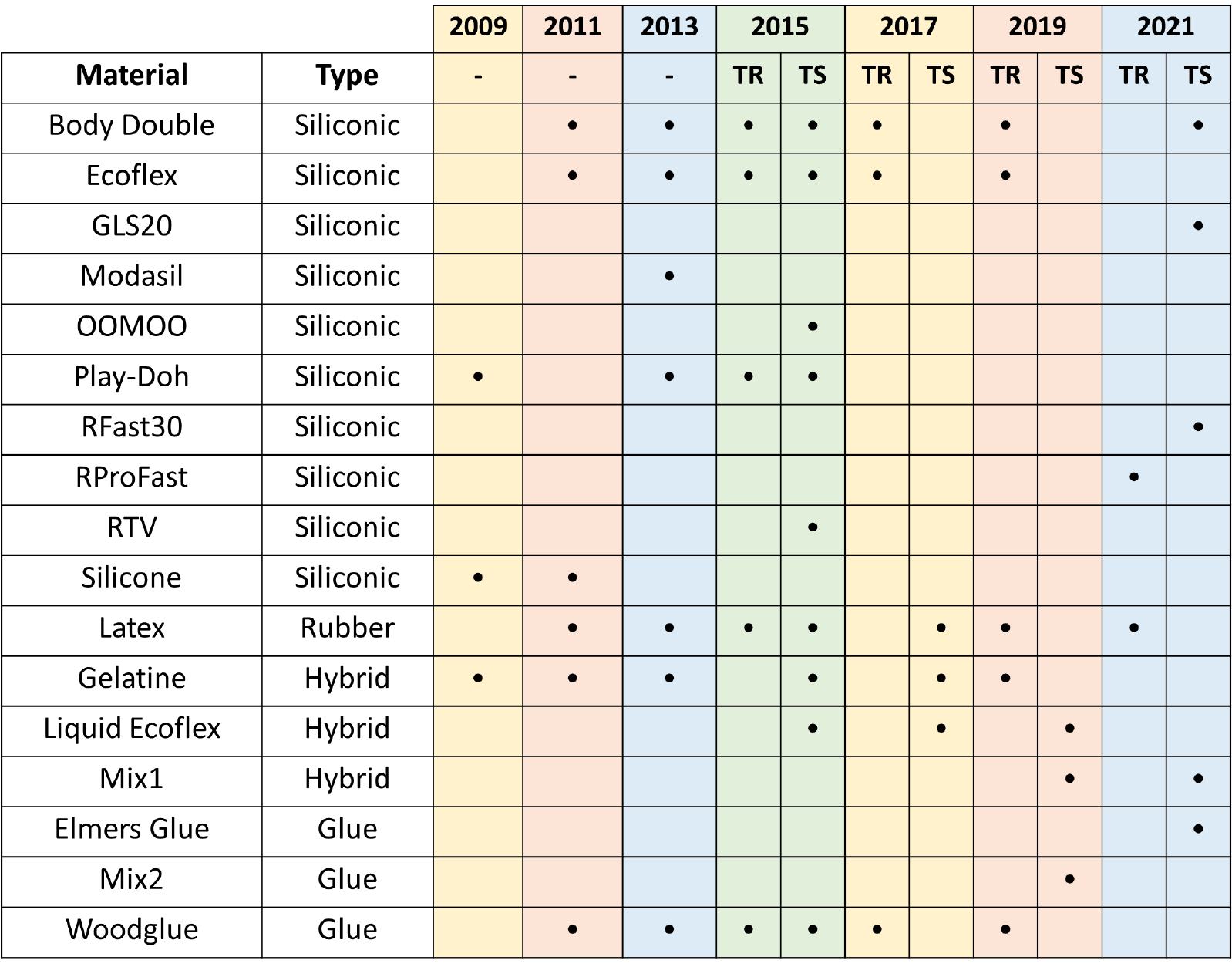}
    \caption{Materials characteristics and frequency over the seven LivDet editions. The train and test materials were completely separeted from 2017 to examine the PADs' resilience against ``never-seen-before'' materials.}
    \label{fig:materials}
\end{figure}

This section will analyze the composition and the methodology for creating each dataset of all the Algorithms part competitions from 2009 to 2021. In these seven editions, a total of twenty-five datasets were created, using fourteen distinct scanners, the majority of which were optical kind, as shown in Table \ref{table:sensorsLivdet}; numerous materials have been tested throughout the years in order to determine the best ones for fabricating fingerprint replicas.
In our experience, the ease with which a successful spoof can be produced depends on subjective factors such as casting the material on the mold or the difficulty of detaching the replica from the mold without ruining one or both, and objective characteristics of composition and behavior. As a matter of fact, certain materials have been shown to be less suited than others. Several of them cannot duplicate the ridges and valleys pattern without introducing visible artifacts such as bubbles and alterations of ridge edges. 
Fig. \ref{fig:materials} reports the best and most “potentially harmful” materials employed in each competition. The final choice was made based on the best trade-off between spoof effectiveness and the fabrication process.

\begin{table}
\caption{Device characteristics for all LivDet datasets. The only kinds of scanner employed in all competitions are optical (O) and thermal swipe (TS).}
\label{table:sensorsLivdet}
\centering
\begin{tabular}{|c|c|c|c|c|c|} 
\hline
\textbf{Scanner}                 & \textbf{Model}  & \textbf{Res. [dpi]} & \textbf{Image size [px]} & \textbf{Type} & \textbf{Edition}  \\ 
\hline
\multirow{2}{*}{Biometrika}      & FX2000          & 569                 & 315x372                  & O             & 2009,2011         \\ 
\cline{2-6}
                                 & HiScan-PRO      & 1000                & 1000x1000                & O             & 2015              \\ 
\hline
\multirow{2}{*}{Crossmatch}      & Verifier 300 LC & 500                 & 480x640                  & O             & 2009              \\ 
\cline{2-6}
                                 & L Scan Guardian & 500                 & 800x750                  & O             & 2013,2015         \\ 
\hline
Dermalog                         & LF10            & 500                 & 500x500                  & O             & 2021              \\ 
\hline
\multirow{2}{*}{DigitalPersona~} & U.are.U 5160    & 500                 & 252x324                  & O             & 2015,2017,2019    \\ 
\cline{2-6}
                                 & 4000B           & 500                 & 355x391                  & O             & 2011              \\ 
\hline
\multirow{2}{*}{GreenBit}        & DactyScan26     & 500                 & 500x500                  & O             & 2015              \\ 
\cline{2-6}
                                 & DactyScan84C    & 500                 & 500x500                  & O             & 2017,2019,2021    \\ 
\hline
Identix                          & DFR2100         & 686                 & 720x720                  & O             & 2009              \\ 
\hline
ItalData                         & ET10            & 500                 & 640x480                  & O             & 2011,2013         \\ 
\hline
Orcanthus                        & Certis2 Image   & 500                 & 300xN                    & TS            & 2017,2019         \\ 
\hline
Sagem                            & MSO300          & 500                 & 352x384                  & O             & 2011              \\ 
\hline
Swipe                            & -               & 96                  & 208x1500                 & TS            & 2013              \\
\hline
\end{tabular}
\end{table}

\begin{figure}[h]
  \centering
  \subfigure[GEL]{\includegraphics[width=0.10\textwidth]{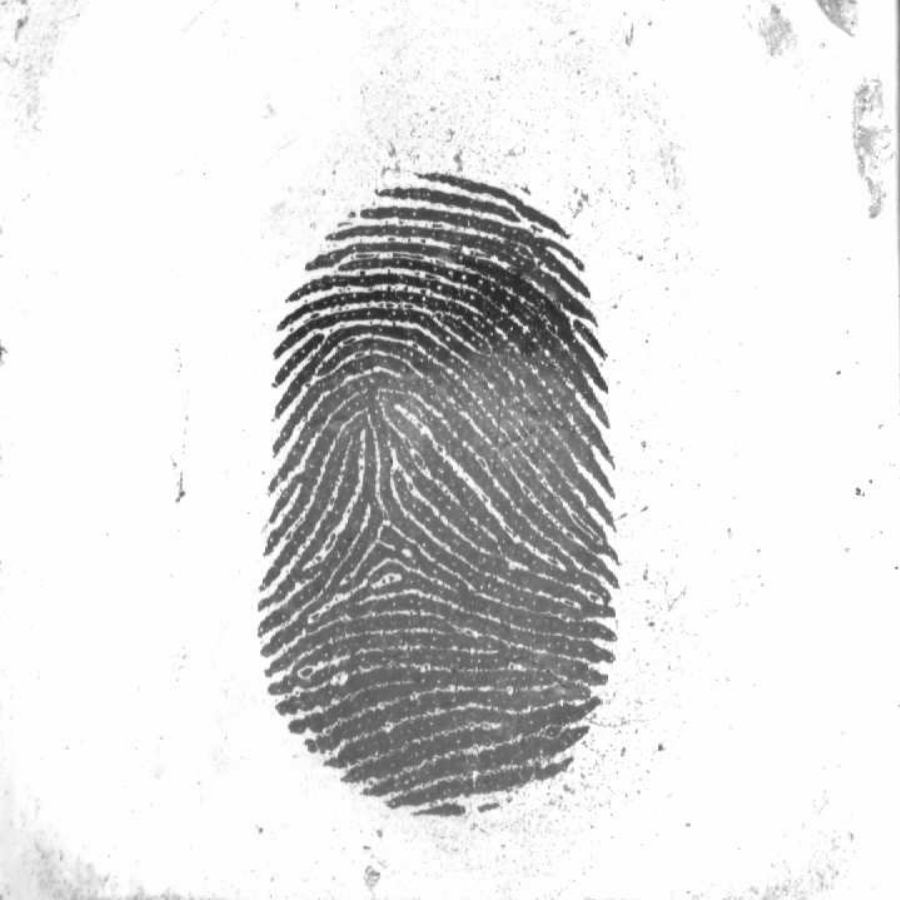}\label{fig:2009gel}}
  \subfigure[LAT]{\includegraphics[width=0.11\textwidth]{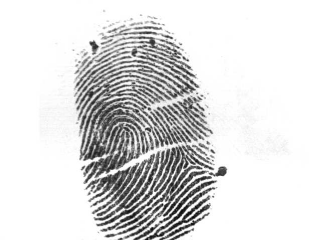}\label{fig:2009lat}}
  \subfigure[SIL]{\includegraphics[width=0.11\textwidth]{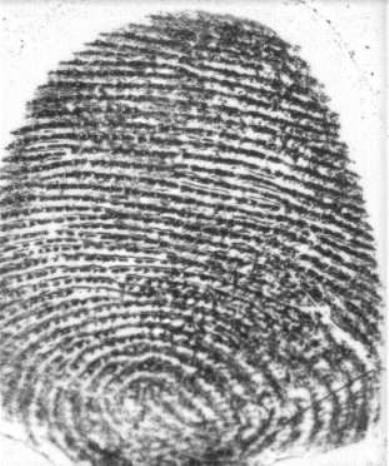}\label{fig:2009sil}}
  \subfigure[LAT]{\includegraphics[width=0.10\textwidth]{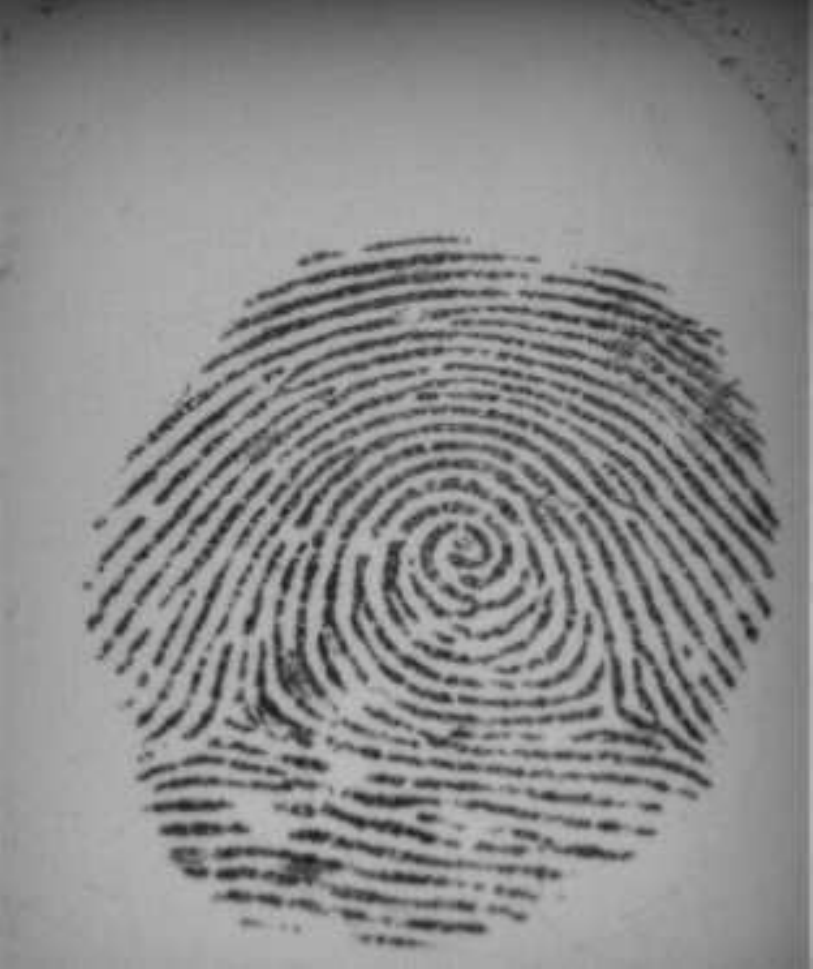}\label{fig:2011latex}}
  \subfigure[PD]{\includegraphics[width=0.11\textwidth]{graphics/eps/2011_pd-eps-converted-to.pdf}\label{fig:2011pd}}
  \subfigure[WG]{\includegraphics[width=0.11\textwidth]{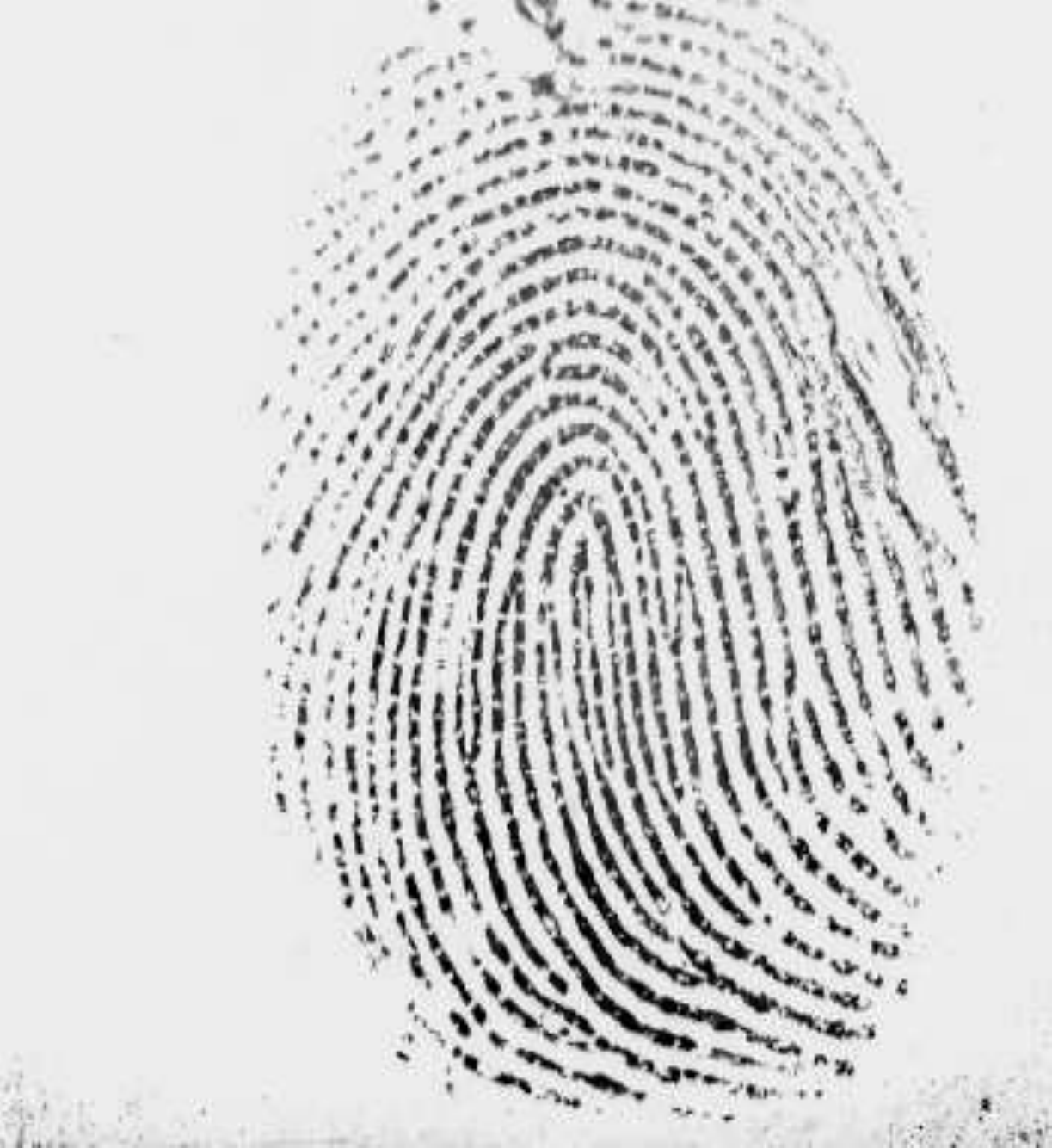}\label{fig:2011wg}}
  \subfigure[BD]{\includegraphics[width=0.10\textwidth]{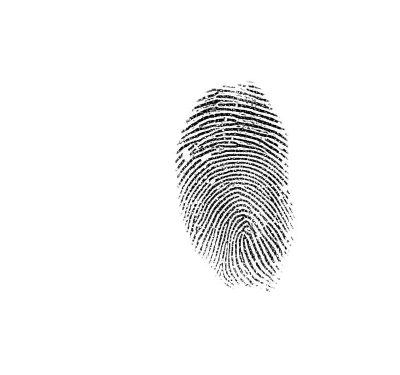}\label{fig:2013bd}}\\
  \subfigure[EC00]{\includegraphics[width=0.11\textwidth]{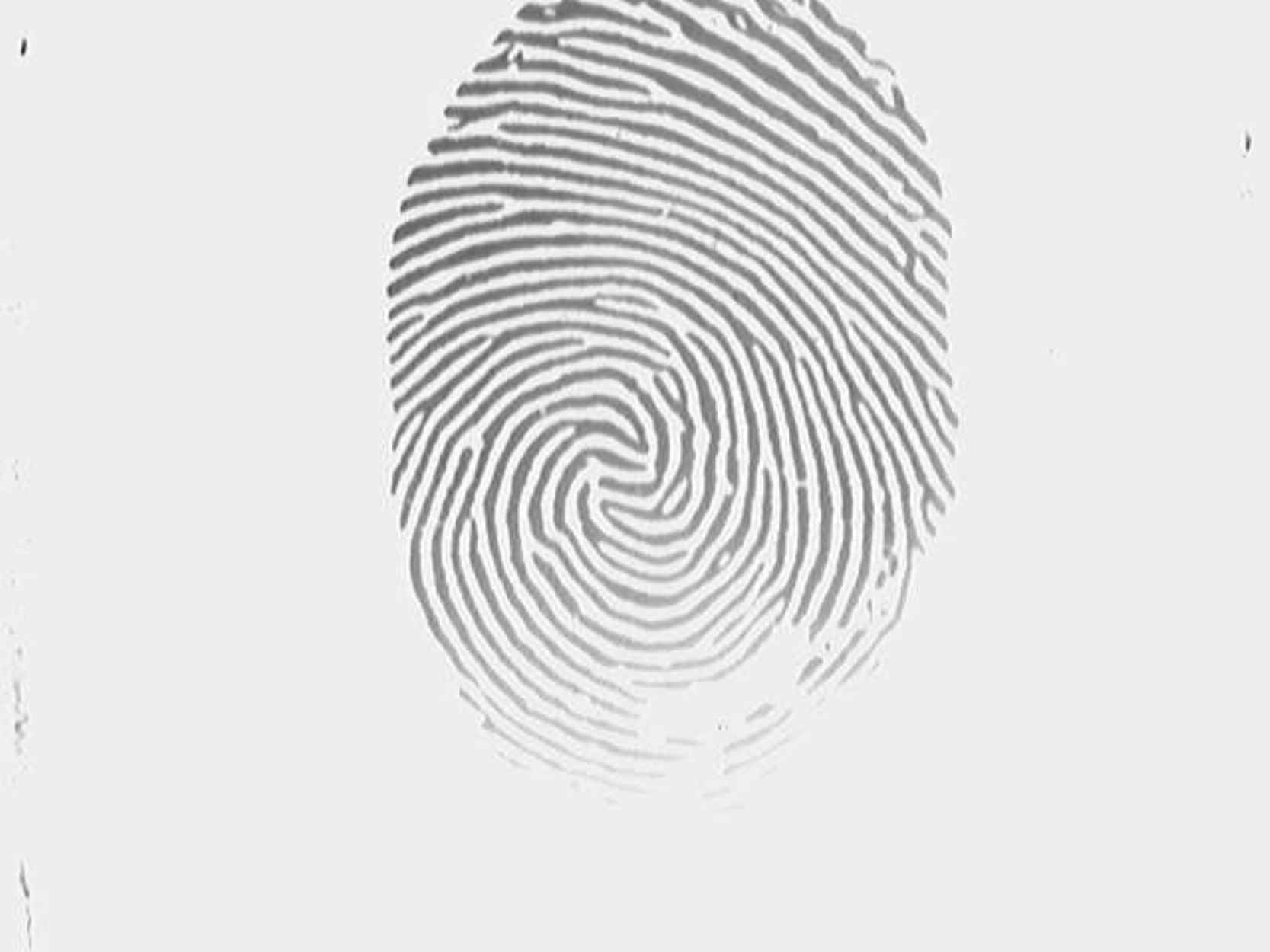}\label{fig:2013ec}}
  \subfigure[MOD]{\includegraphics[width=0.11\textwidth]{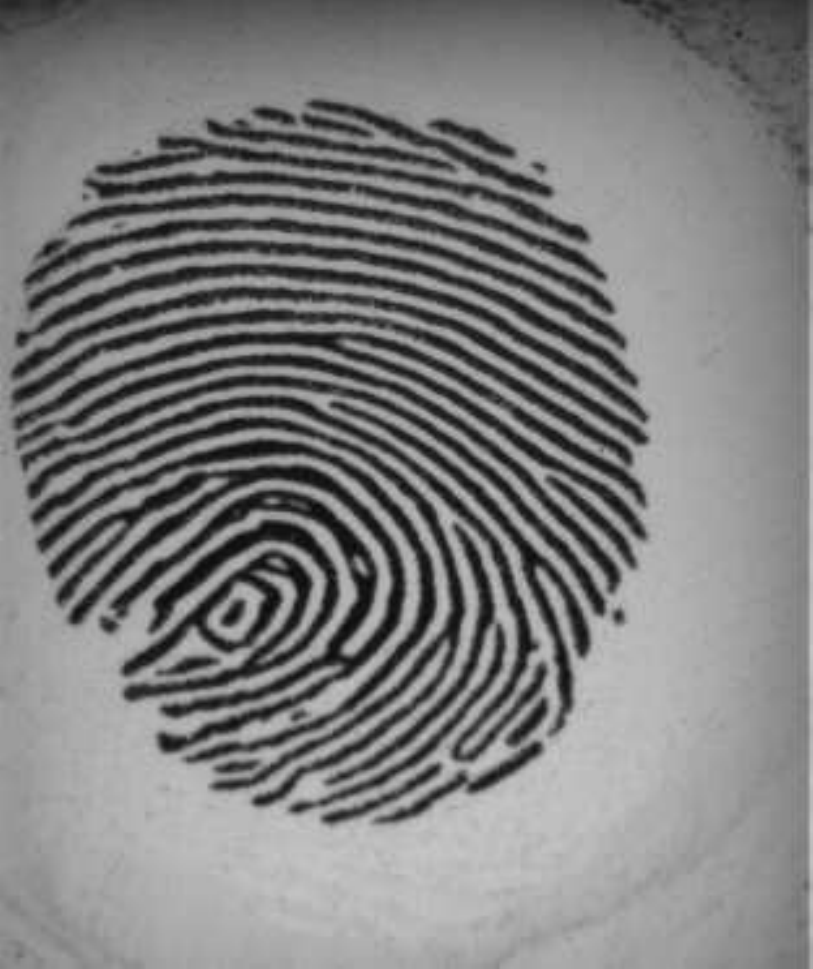}\label{fig:2013mod}}
  \subfigure[EC00]{\includegraphics[width=0.10\textwidth]{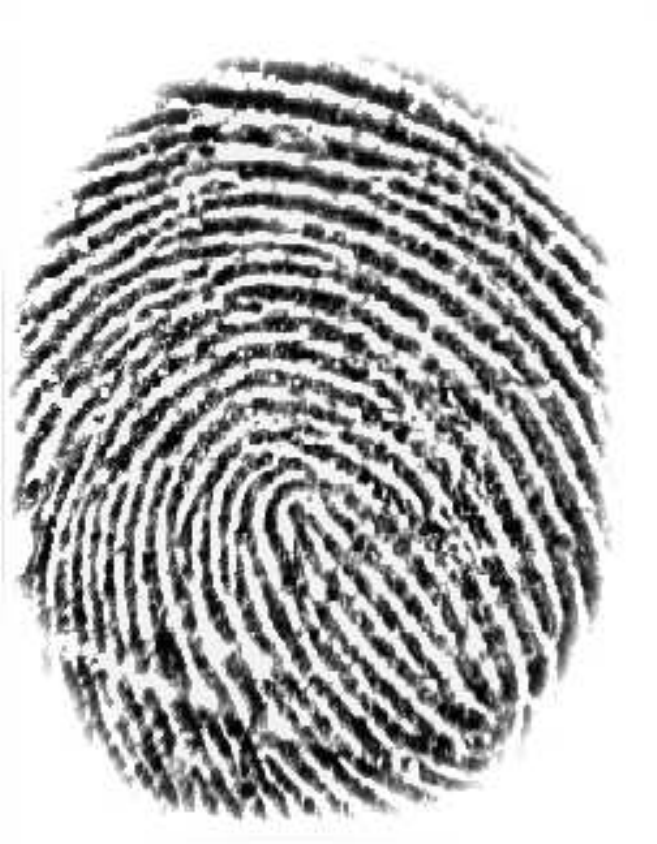}\label{fig:2015ec00}}
  \subfigure[LAT]{\includegraphics[width=0.11\textwidth]{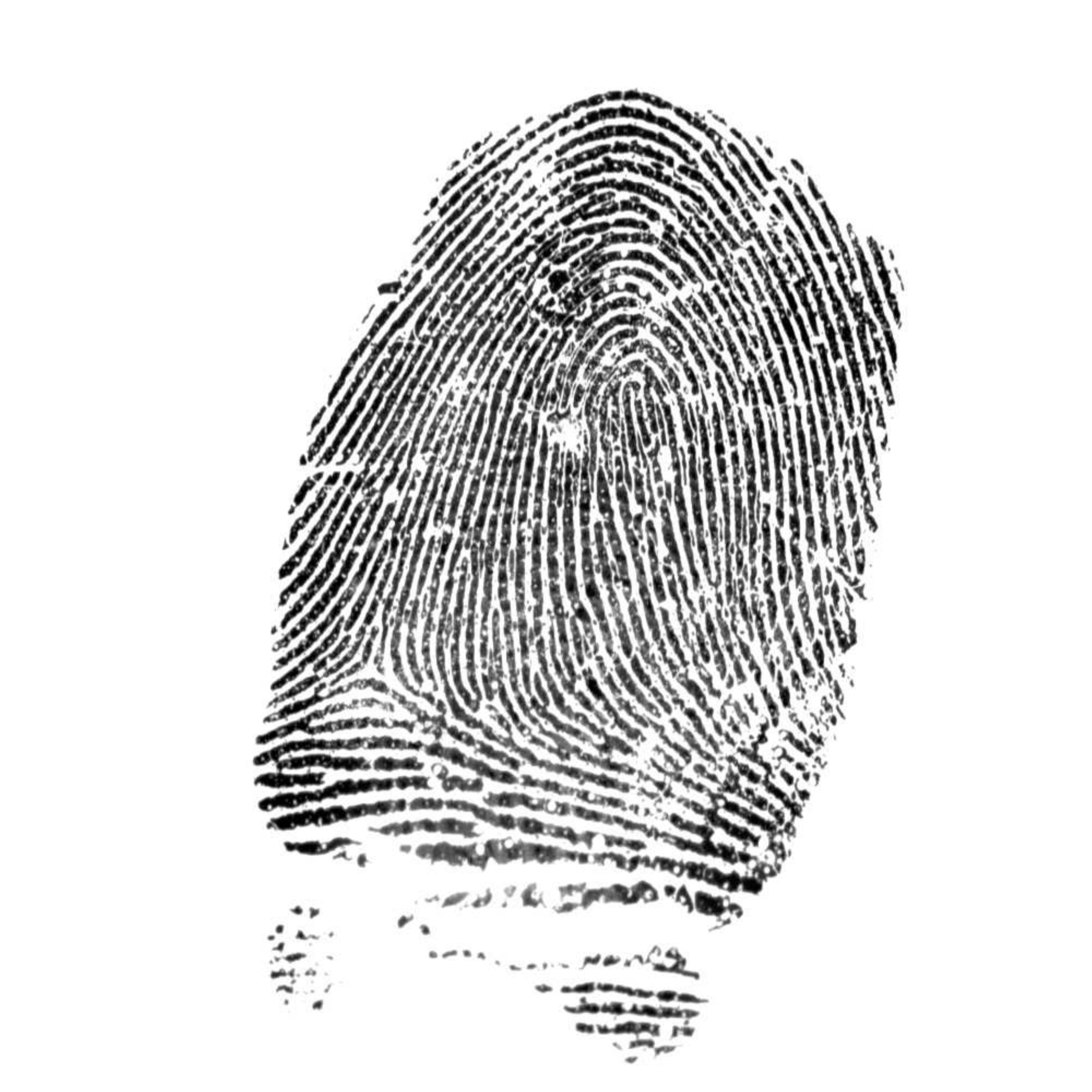}\label{fig:2015lat}}
  \subfigure[LE]{\includegraphics[width=0.11\textwidth]{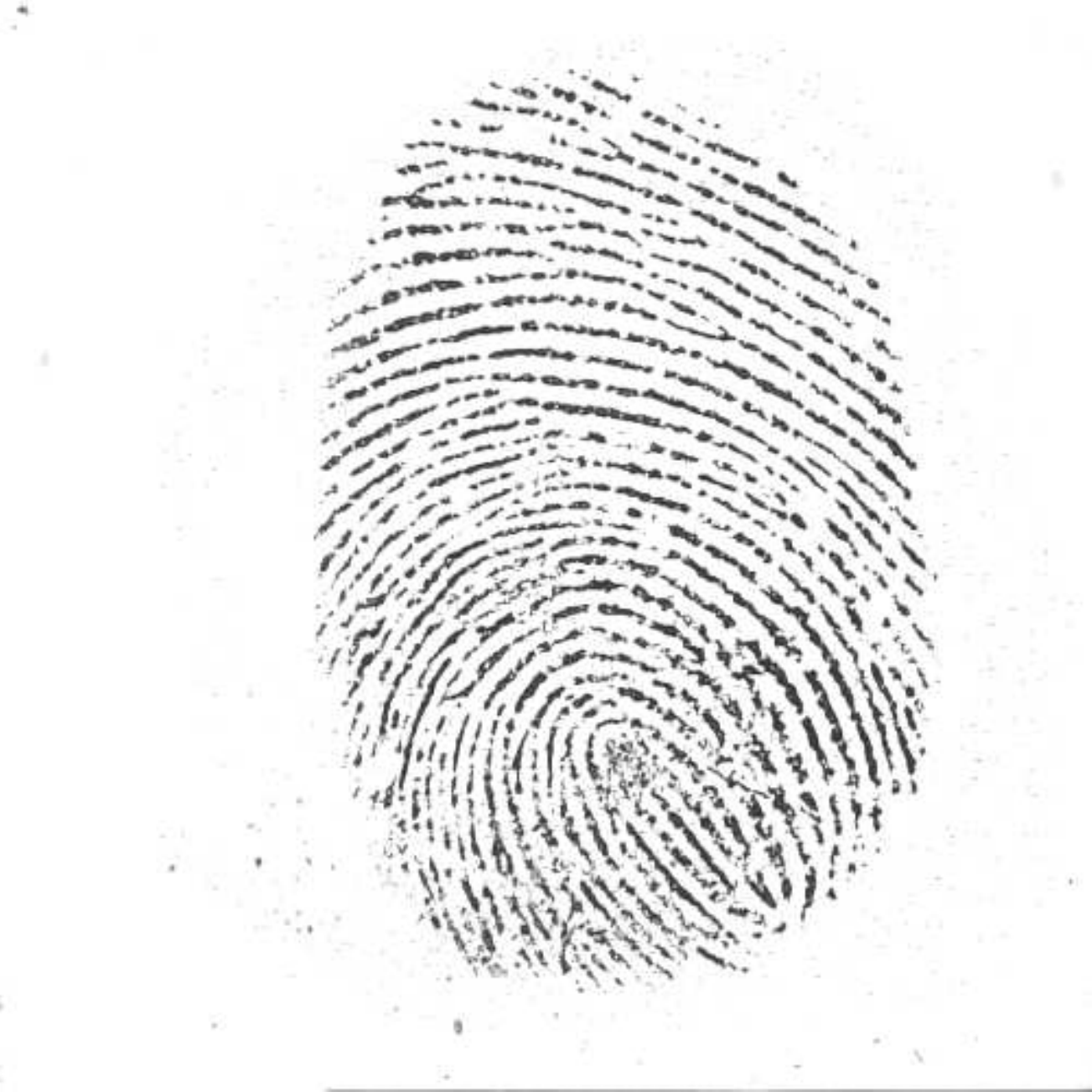}\label{fig:2015le}}
  \subfigure[BD]{\includegraphics[width=0.10\textwidth]{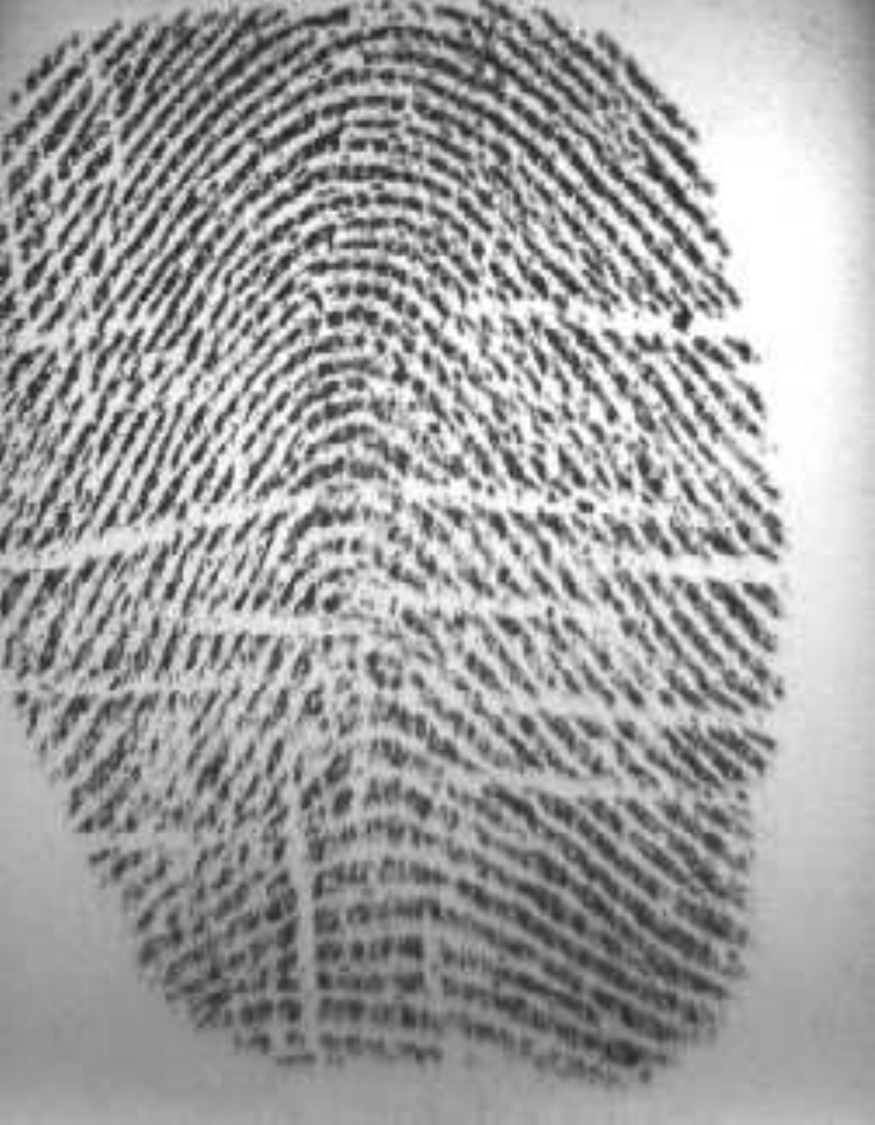}\label{fig:2017bd}}
  \subfigure[LAT]{\includegraphics[width=0.11\textwidth]{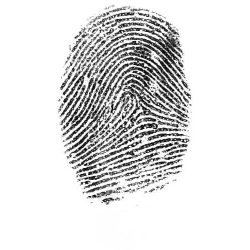}\label{fig:2017lat}}\\
  \subfigure[WG]{\includegraphics[width=0.11\textwidth]{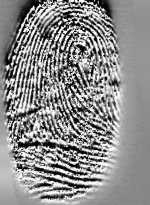}\label{fig:2017wg}}
  \subfigure[LE]{\includegraphics[width=0.10\textwidth]{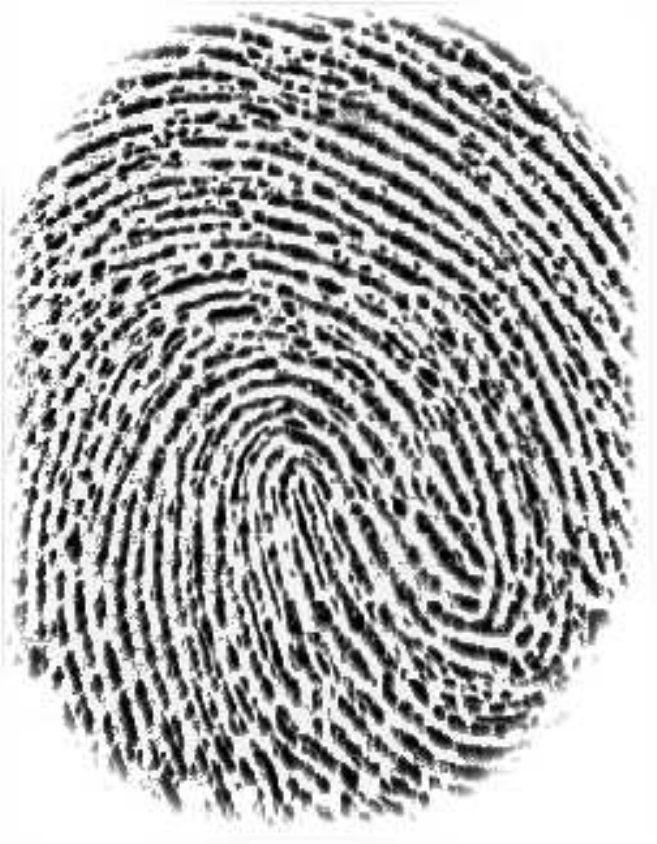}\label{fig:2019le}}
  \subfigure[Mix1]{\includegraphics[width=0.11\textwidth]{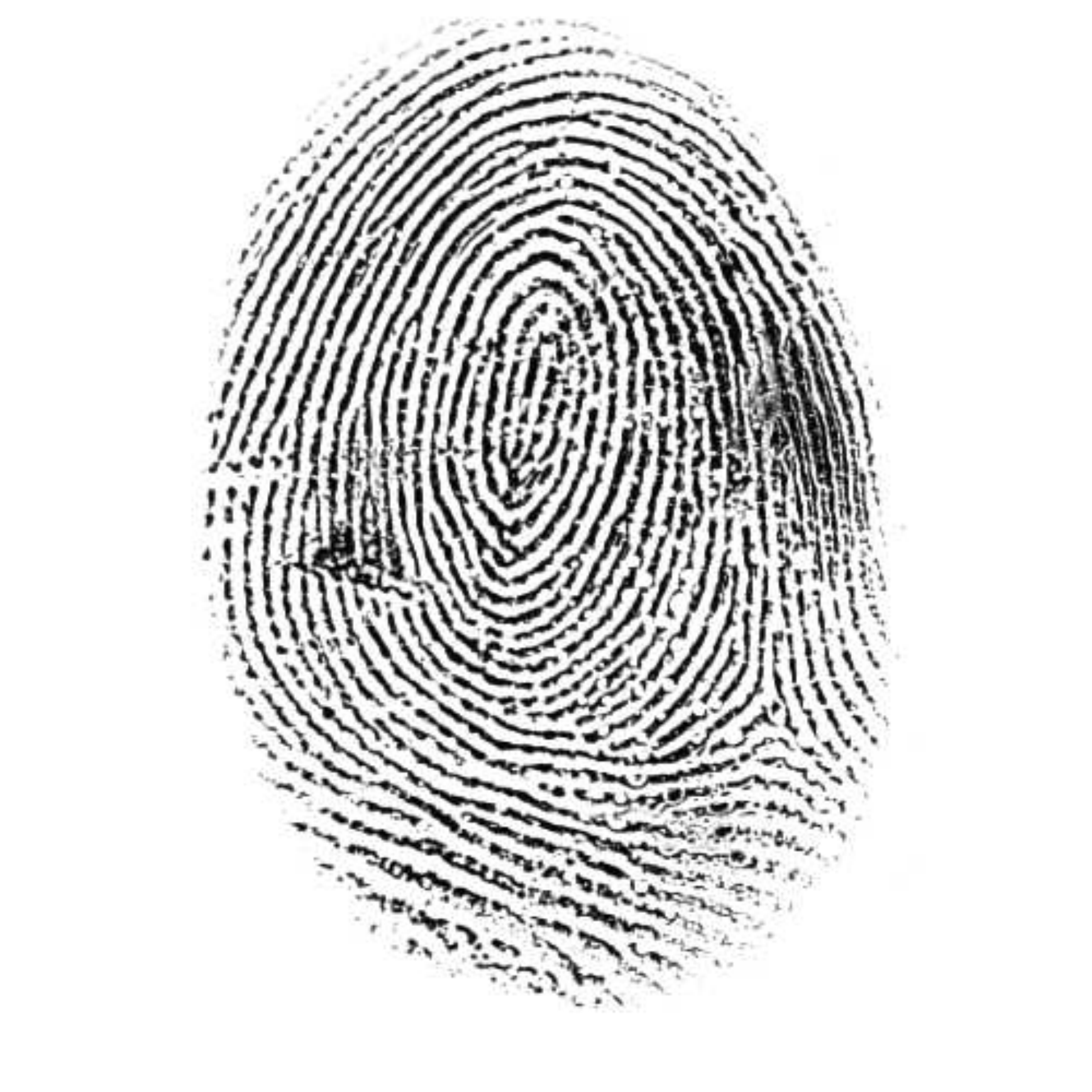}\label{fig:2019mix1}}
  \subfigure[Mix2]{\includegraphics[width=0.11\textwidth]{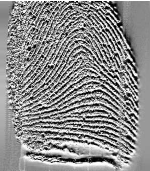}\label{fig:2019mix2}}
  \subfigure[BD]{\includegraphics[width=0.10\textwidth]{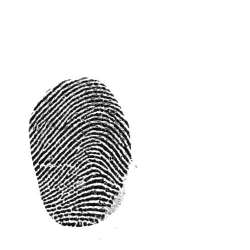}\label{fig:2021bd}}
  \subfigure[GLS]{\includegraphics[width=0.11\textwidth]{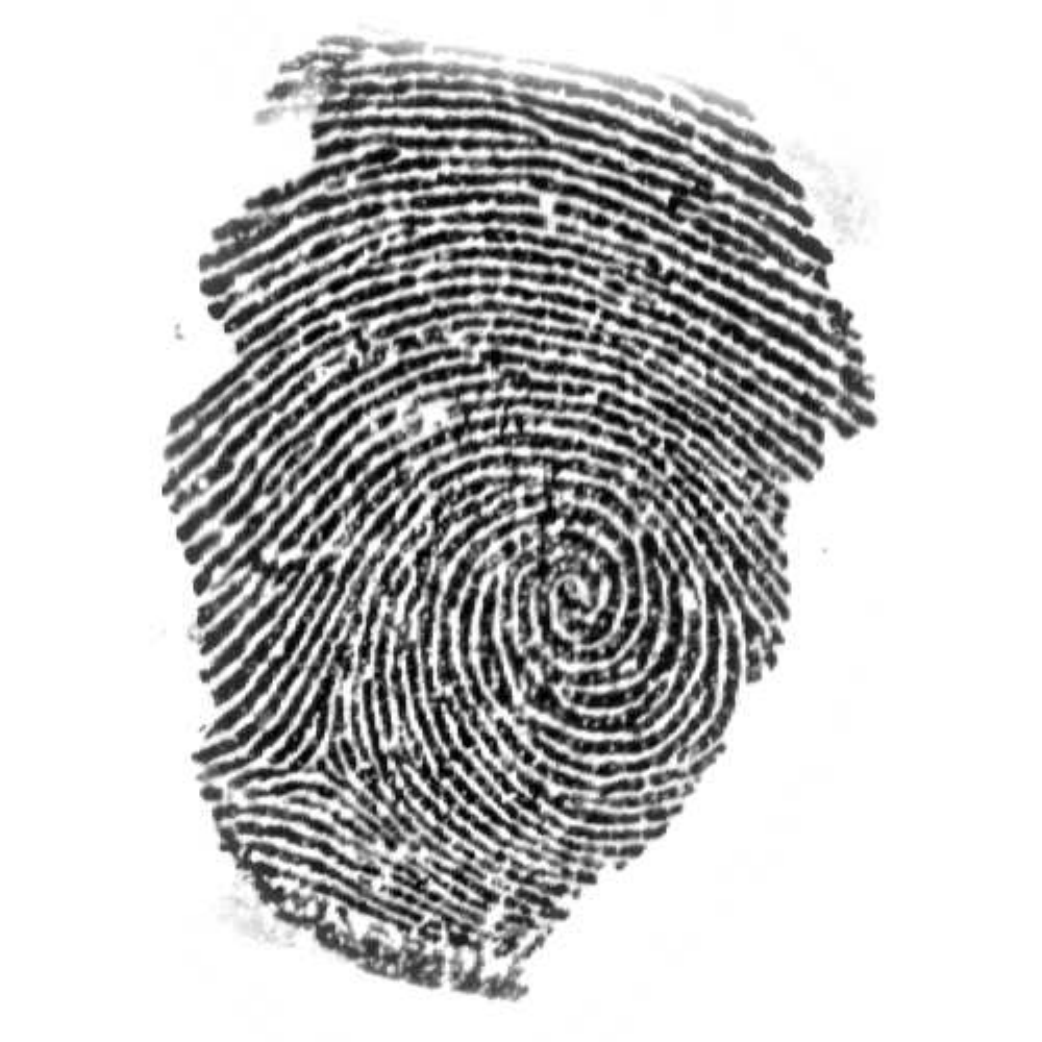}\label{fig:2021gls}}
  \subfigure[RF30]{\includegraphics[width=0.11\textwidth]{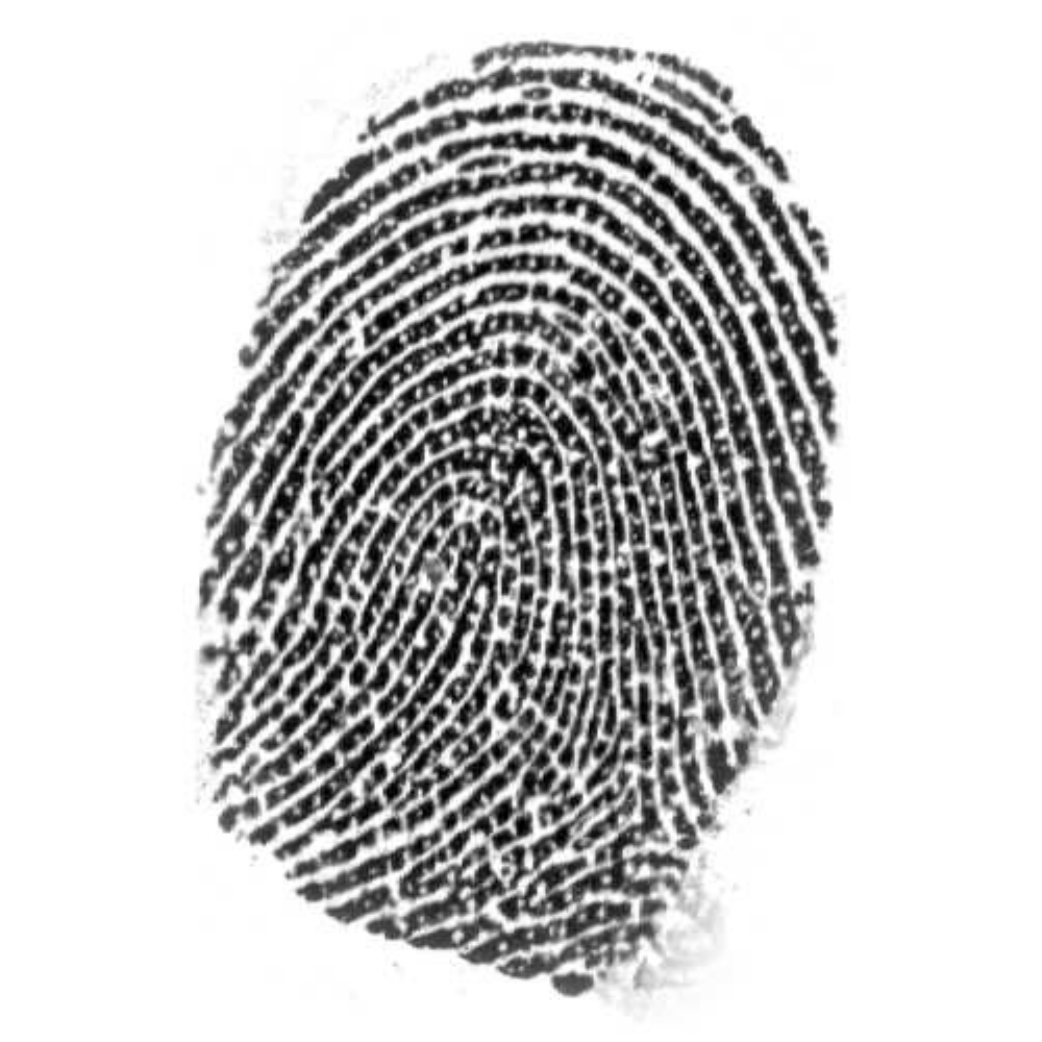}\label{fig:2021rfast}}
  \caption{Examples of successful spoof attacks with various materials.}  \label{fig:allfake}
\end{figure}

In the first edition of 2009, the three optical scanners Crossmatch, Identix, and Biometrika were used. The fingerprint images are created via the consensual method, utilizing gelatine, silicone, and play-doh as spoof materials (Fig. \ref{fig:allfake}\{a,c\}).

LivDet 2011 presents four datasets related to four optical sensors: Biometrika, Digital Persona, ItalData, and Sagem. Consensual spoofs were fabricated with gelatin, latex, ecoflex, play-doh, silicone, and woodglue (Fig. \ref{fig:allfake}\{d,f\}).

The 2013 dataset consists of images from four devices: Biometrika, Crossmatch, Italdata, and Swipe, the first non-optical scanner introduced in the competition. However, this edition stands out for another relevant novelty:  the non-cooperative method in producing the spoof images of the Biometrika and Italdata datasets. In contrast, the Crossmatch and Swipe datasets were created with the consensual approach. Except for silicone, the adopted materials are the same as the previous edition, with the addition of body double and modasil (Fig. \ref{fig:allfake}\{g,i\}). 

LivDet 2015 is again characterized by four optical scanners: GreenBit, Biometrika, Digital Persona, and Crossmatch. The fake fingerprints were all created through the cooperative technique. 
For the Green Bit, Biometrika, and Digital Persona datasets, spoof materials included ecoflex, gelatine, latex, woodglue, a liquid ecoflex, and RTV (a two-component silicone rubber); for the Crossmatch dataset, play-doh, Body Double, Ecoflex, OOMOO (a silicone rubber), and a novel form of gelatin (Fig. \ref{fig:allfake}\{j,l\}). In this edition, the testing sets included spoof images of unknown materials, i.e. materials which were not included in the training set. This strategy has been used to determine the robustness of algorithms in facing unknown materials.
The results obtained prompted the organizers to mantain the same protocol for subsequent editions. 

LivDet 2017 consisted of data from three fingerprint sensors: Green Bit DactyScan84C, Digital Persona U.are.U 5160 configured on an Android device, and a new thermal swipe sensor called Orcanthus.
The materials adopted to generate the training set are wood glue, ecoflex, body double, whereas gelatine, latex and liquid ecoflex compose the test set, namely, it is wholly made up of never-seen-before materials;(Fig. \ref{fig:allfake}\{m,o\}). 
It's worth mentioning that this LivDet version also features a new characteristics: two sets of persons with varying degrees of ability were involved in manufacturing spoofs and gathering associated images.
In this manner, we may simulate a real-world scenario in which the operator's capacity to build the train set differs from that of potential attackers. 

The scanners in LivDet 2019 are the same as in previous editions, albeit all three were tested on Windows OS. This edition presents for the first time multi-material combinations, both of different consistency and of different nature. The others were the same of the previous edition (Fig. \ref{fig:allfake}\{p,r\}). 

The latest edition features only two scanners, GreenBit and Dermalog. For each of them, two datasets were created: the first through the consensual approach and the second through the new pseudo-consensual method, called ScreenSpoof \cite{casula2021spoofs}. Except for latex, body double, and ``mix1'' (already present in previous competitions), new materials have been chosen for the fakes, such as RProFast, Elmers glue, gls20, and RPro30 (Fig. \ref{fig:allfake}\{s,u\}). 

\subsubsection{Specific Challenges}

With the passing of the editions, the organizers have made several innovations, both in terms of the production of fakes and in terms of the difficulties that participants must address: LivDet 2013 \cite{livdet2013} introduced a dataset with spoofs created by latent fingerprints; LivDet 2015 \cite{livdet2015} included some ``unknown'' materials within the dataset used for detectors' evaluation; in LivDet 2017 \cite{livdet2017}, train and test sets have been collected by different skilled operators, and they were not completely separated since some users are present in both of them, in order to test the so-called ``user-specific'' effect \cite{ghiani2016user}; in LivDet 2019 \cite{livdet2019} participants were asked to submit a complete algorithm that could output not only the image's liveness probability, but also an integrated score which included the probability of fingerprint matching, and finally LivDet 2021 \cite{livdet2021} presented two datasets created using the new and effective ScreenSpoof method. The most significant novelties introduced in the competitions will be discussed thoroughly in the following Sections.

\subsubsection*{LivDet 2013 - Consensual and semi-consensual}

In the cooperative artifact fabrication scenario, the subject is requested to press his finger on a moldable and stable substance to get the negative of his fingerprint. The actual fingerprint can be obtained by casting the spoof material on the negative mold.
This method is usually considered the ``worst-case'' scenario since it allows to fabricate high-quality spoofs, although this type of attack has a low probability of being carried out since it requires the victim's collaboration.
On the other hand, semi-consensual spoof generation requires the volunteer to leave his fingerprint on a smooth or non-porous surface. The latent fingerprint is then highlighted by applying magnetic powders so that it can be photographed or scanned. The scanned image is then enhanced through image processing and finally printed on a transparent sheet from which the spoof is generated.
In this case, the development stage is a destructive operation, since it can only be done once per trace; therefore, the quality of the mold is entirely dependent on the attacker's ability to complete it without compromising the latent mark's quality. 
Furthermore, the scenario in which the attacker uses forensic techniques to generate and lift the victim's latent fingerprint is not feasible without the attacker having considerable technical expertise and time.

In the following sections, we will analyze how the systems' performance and the fakes' quality vary as the spoofing method used varies.

\subsubsection*{LivDet 2015 - Hidden materials} 

The materials used to create the fakes in the first three editions of the competition were essentially the same. This constant may have led competitors to enter parameters capable of recognizing the characteristics of a specific material and, therefore, of easily distinguishing a live fingerprint from a fake. To try to manage any type of material, some materials never used before were included in the LivDet 2015 \cite{livdet2015} test set: liquid ecoflex and RTV for the Green Bit, Biometrika, and Digital Persona sensors OOMOO and a new gelatine for Crossmatch. As a matter of fact, in a realistic scenario, it is not possible to determine whether the material used for the attack is present in the dataset used to train the system. Therefore, the competitors' results explain not only the system's accuracy but also its cross-material reliability. This novelty was maintained for all subsequent editions: from 2017 onwards the train and test materials were different to evaluate the robustness of the PADs to the so-called ``never-seen-before'' materials.


\subsubsection*{LivDet 2017 - User-specific effect and operator skill}

The 2017 \cite{livdet2017} edition of Livdet was focused on determining how much the data composition and collection techniques influence the PAD systems. Although the aim of the competition was clear from the beginning, the following aspects have been analyzed in a later work \cite{orru2019analysis} and only marginally in the LivDet 2017 report for the sake of space: (1) the presence of operators with different skills in fabricating the replicas, one with a considerable degree of expertise (high skilled) and one composed of novice forgers (low skilled)
; (2) the presence of some users in both the training and testing parts of the three datasets. As a matter of fact, in previous editions, the train and test sets were completely separated as none of the users were present in both. Nevertheless, it has been demonstrated \cite{ghiani2016user} that having different acquisitions of the same fingerprint in both parts of the dataset resulted in substantially higher classification results when compared to a system with distinct users in train and test.
The additional information coming from the enrolled user was called ``user-specific effect'', whereas PAD systems are often based on ``generic user'' techniques because the final user population is unknowable.
User-specific information is essential when creating a fingerprint presentation attacks detector that must be incorporated into a fingerprint verification system. It is assumed that the system will have some knowledge of the users in this scenario, which may be used to develop a more robust software module to presentation attacks when authenticating subjects. It is worth noting how the focus was starting to move from stand-alone PAD to integrated systems in this edition. This tendency will lead to the next edition ``LivDet in action'' challenge.

\subsubsection*{LivDet 2019 - LivDet in Action}

In the early 2000s, most competitions focused on the matching process, such as the Fingerprint Verification Competition \cite{cappelli2007fingerprint}.

Since its inception, the Liveness Detection Competition series has been running to provide a standard for assessing PAD algorithms in the same way that matching performance was measured.
The subsequent LivDet editions were therefore focused on evaluating the performance of a standalone fingerprint liveness detector. 
Nevertheless, in real-world applications, the FPAD system is generally used in conjunction with a recognition system to determine whether the input is from a genuine user or an impostor attempt to deceive the system \cite{MarcelInAction}.
For this purpose, in LivDet 2019 \cite{livdet2019}, we invited all the competitors to submit a complete algorithm capable of outcoming not only the classic image's liveness probability but also an integrated match score (IMS) that includes the degree of similarity between the given sample and the reference template.
Based on state-of-the-art methods in such fields, our investigation provided a clear overview of how integrating a standard verification system with a PAD may affect final performances.
This challenge, considering its importance in a real application context, was also maintained for the next LivDet2021 edition.

\subsubsection*{LivDet 2021 - ScreenSpoof method}

In comparison to previous editions, the latest represents a turning point. Firstly, two distinct challenges were designed to test the competitor algorithms: the first one, called ``Liveness Detection in Action'' in the wake of the previous edition, investigated the impact of combining a PAD system with a fingerprint verification system; the second task, named ``Fingerprint representation'', was designed to test the compactness and representativeness achieved by feature sets adopted, which are significant factors to ensure reliable performances in terms of accuracy and speed in current biometric systems.
Moreover, in this edition, two different methods for creating spoofs were used, as in LivDet 2013. We took inspiration from our recent work \cite{casula2021spoofs}, where we have shown that it is possible to create fine-quality replicas by properly pre-processing the snapshots of the latent fingerprints left on the smartphone screen surface. We called this technique ``ScreenSpoof''. Although it is a semi-consensual fabrication method, our findings revealed a threat level comparable to that of attacks using spoofs fabricated with the complete consensus of the victim.
For this reason, this acquisition protocol was repeated in the last edition.
Unlike LivDet 2013, the results obtained for this new method are as effective as the consensual method and, in some specific cases, even more successful.


\subsection{LivDet Systems}

The LivDet Systems competition does not produce the publically available datasets seen in the Algorithms competition; however, with each competition, the dataset collected on a submitted system is supplied to the respective competitors to assist them in improving their systems. 
Each submission includes a fully packaged fingerprint system with software installation. The systems were equipped with only a liveness module until the 2017 edition \cite{8698578}, in which we also accepted integrated systems, with enrollment, matching and liveness modules.
A system outputs a score between 0-100 representing liveness or 0-100 for match scores. In particular, in the PAD-only competition, a score greater than 50 indicates a live sample, while in the integrated systems competition, a score greater than 50 means that a sample is both a match and live. Each competition contained several PAIs, with a portion of them known to the competitor ahead of time and the remainder unknown. The 2017 systems competition also included the concept of unknown mold types with the use of 3D printed molds \cite{8267123} for some PAIS being kept from the competitors.

\subsection{Performance evaluation}

Another significant step forward of the latest edition regards the nomenclature adopted for the performance evaluation, which has been updated in favor of ISO/IEC 30107-1:2016 terminology \cite{iso}. 
Table \ref{table:nomenclature} reports the related terms and their meaning. In particular, APCER and BPCER evaluate the performance of the FPADs, whereas FNMR, FMR and IAPMR assess the performance of an integrated system.
Additional metrics often employed and deriving from those previously mentioned are:
\begin{itemize}
    \item \textit{Liveness Accuracy}: percentages of samples correctly classified by the PAD, that is the inverse of the weighted average of APCER and BPCER.
    \item \textit{Integrated Matching (IM) Accuracy}: percentages of samples correctly classified by the integrated system, that is the inverse of the weighted average of FNMR, FMR and IAPMR.
\end{itemize}

\begin{table}
\centering 
\caption{Correspondence between the ISO terms and the terms used in older Livdet editions (2009-2019).}
\resizebox{1\textwidth}{!}{%
\begin{tabular}{|l|l|l|} 
\hline
\textbf{ISO/IEC 30107-1 term}                            & \textbf{Old LivDet term} & \textbf{Meaning}                                                                    \\ 
\hline
\begin{tabular}[c]{@{}l@{}}Attack Presentation Classification \\ Error Rate (APCER)\end{tabular}  & FerrFake     & Rate of misclassified fake~fingerprints.  \\ 
\hline
\begin{tabular}[c]{@{}l@{}}Bona fide Presentation Classification \\ Error Rate (BPCER)\end{tabular} & FerrLive                 & Rate of misclassified live fingerprints.                                            \\ 
\hline
 False non-match rate (FNMR)                           & \begin{tabular}[c]{@{}l@{}}100-IMG\_accuracy\\100-Genuine Acceptance Rate (GAR)\\ False Reject Rate (FRR)\end{tabular} & \begin{tabular}[c]{@{}l@{}}Rate of incorrectly rejected genuine live fingerprints.\end{tabular}                                \\ 
\hline
False match rate  (FMR)                                  &        False Accept Rate (FAR)                  & \begin{tabular}[c]{@{}l@{}}Rate of incorrectly accepted zero-effort impostors\end{tabular}                                                                                \\
\hline
\begin{tabular}[c]{@{}l@{}}Impostor Attack Presentation\\  Match Rate (IAPMR)\end{tabular}          & \begin{tabular}[c]{@{}l@{}}100-IMI\_accuracy\\  100-SGAR\end{tabular}          & Rate of incorrectly classified impostor live or genuine fake fingerprints.             \\
\hline
\end{tabular}}
\label{table:nomenclature}
\end{table}







\section{Examination of results}
\label{sec:results}
Different challenges, different types of PADs, and different materials make the comparison between LivDet editions non-trivial.
The purpose of this review is to compare the editions from different points of view to highlight the differences between acquisition sensors, spoofing techniques, and the nature of PADs.

\begin{table}[]
\caption{Mean and standard deviation BPCER and APCER values for each LivDet competition.}
\label{table:generalcomp}
\resizebox{\textwidth}{!}{%
\begin{tabular}{|l|c|c|c|c|c|c|c|}
\hline
\multicolumn{1}{|c|}{\textbf{}} & \textbf{LivDet 2009} & \textbf{LivDet 2011} & \textbf{LivDet 2013} & \textbf{LivDet 2015} & \textbf{LivDet 2017} & \textbf{LivDet 2019} & \textbf{LivDet 2021} \\ \hline
\textbf{BPCER}                  & 22.03 ($\pm$ 9.94)     & 23.52 ($\pm$ 13.85)    & 6.26 ($\pm$ 9.18)      & 5.94 ($\pm$ 3.21)      & 4,99 ($\pm$ 0.89)      & 4,15 ($\pm$ 2.65)      & 1,1 ($\pm$ 0.65)       \\ \hline
\textbf{APCER}                  & 12.47 ($\pm$ 6.12)     & 26.97 ($\pm$ 18.24)    & 21.55 ($\pm$ 37.21)    & 6.89 ($\pm$ 3.83)      & 4,97 ($\pm$ 0.62)      & 4,75 ($\pm$ 5.71)      & 27,02 ($\pm$ 23.31)    \\ \hline
\end{tabular}}
\end{table}

A first comparison is reported in Table \ref{table:generalcomp} which shows the mean and standard deviation of APCER and BPCER of the three best algorithms of each edition.
Although it is necessary to weigh the differences between edition and edition, there is an evident downward trend in errors in both the classification of live and fake samples.

This trend was interrupted only in the last edition regarding the APCER, which has very high values. This increase is partially explained by the presence of a new forgery methodology, the ScreenSpoof method \cite{casula2021spoofs}. This aspect is discussed in more detail in Section \ref{section:unconsensual}. The APCER values relating to the consensual acquisition method alone are, however, very high: the three best methods obtained an average APCER of 17.13\% on the LivDet 2021 dataset acquired through the consensual technique. Another reason for the drop in performance lies in the materials used to manufacture the spoofs: in fact, the 2021 training set is the only one that does not contain any glue-like material; for this reason, we hypothesize that materials of this nature in the test set are very insidious for PADs. The analysis of the materials of the last three editions is reported in Section \ref{section:materials}.

\begin{figure}
    \centering
    \includegraphics[width=\textwidth]{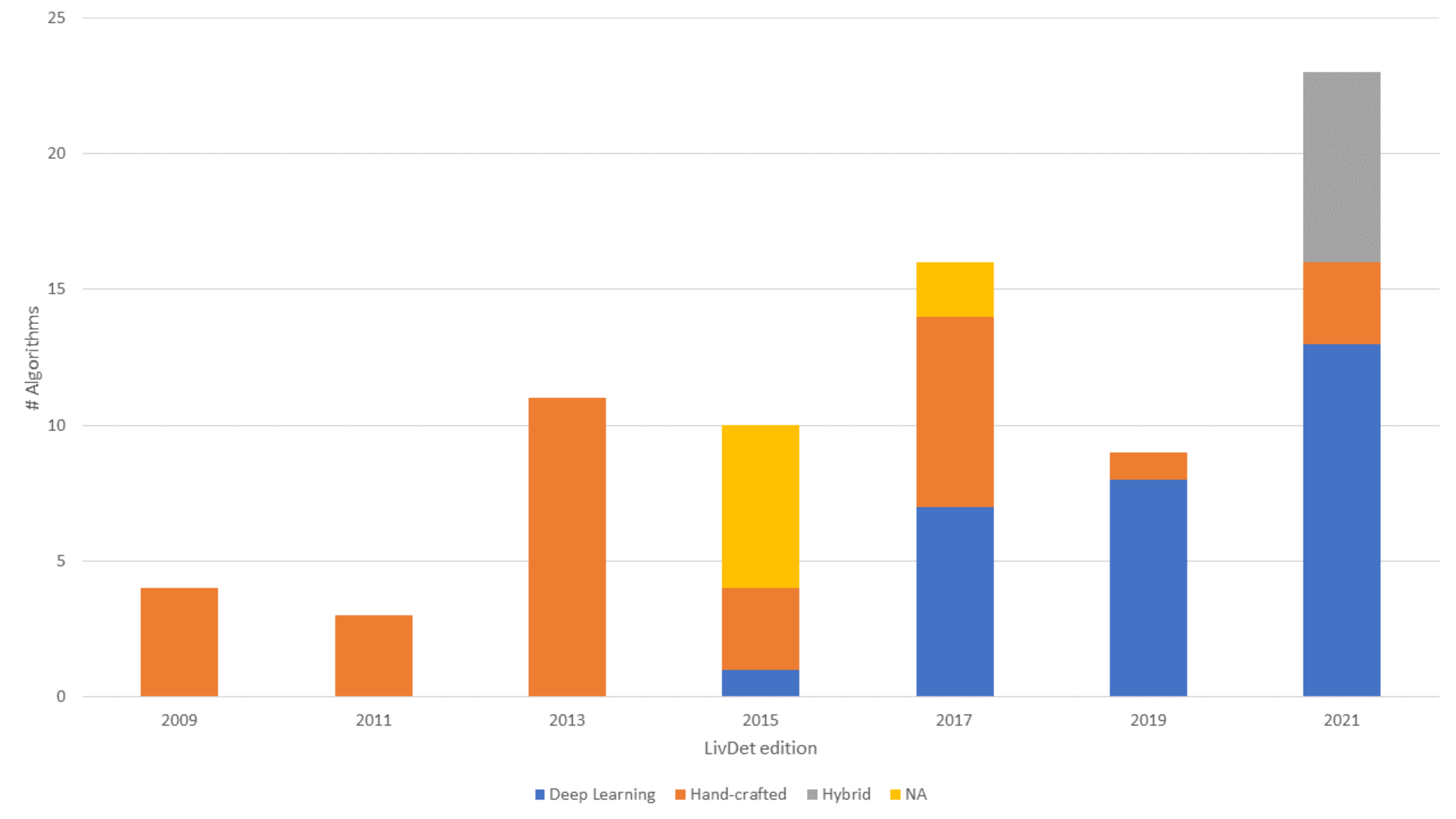}
    \caption{Types of PAD submitted to LivDet - algorithms part.}
    \label{fig:pad_typ}
\end{figure}

The combined examination of the Table \ref{table:generalcomp} and Figure \ref{fig:pad_typ}, showing the types of PADs of the various editions, underlines the contribution of deep learning techniques in liveness detection starting from the 2015 edition, whose winner \cite{7390065} has presented a CNN-based solution. In the following competitions, submissions based on deep-learning have increased, up to become the most numerous since 2019. In the last edition, several hybrid solutions were presented.

It is important to underline that the difficulty of LivDet datasets has increased since 2015, with the introduction of ``never-seen-before'' materials in the test sets. Therefore, the leap in the precision of the PADs from 2013 to 2015 and subsequent editions is to be considered more significant.

\begin{figure}
    \centering
    \includegraphics[width=\textwidth]{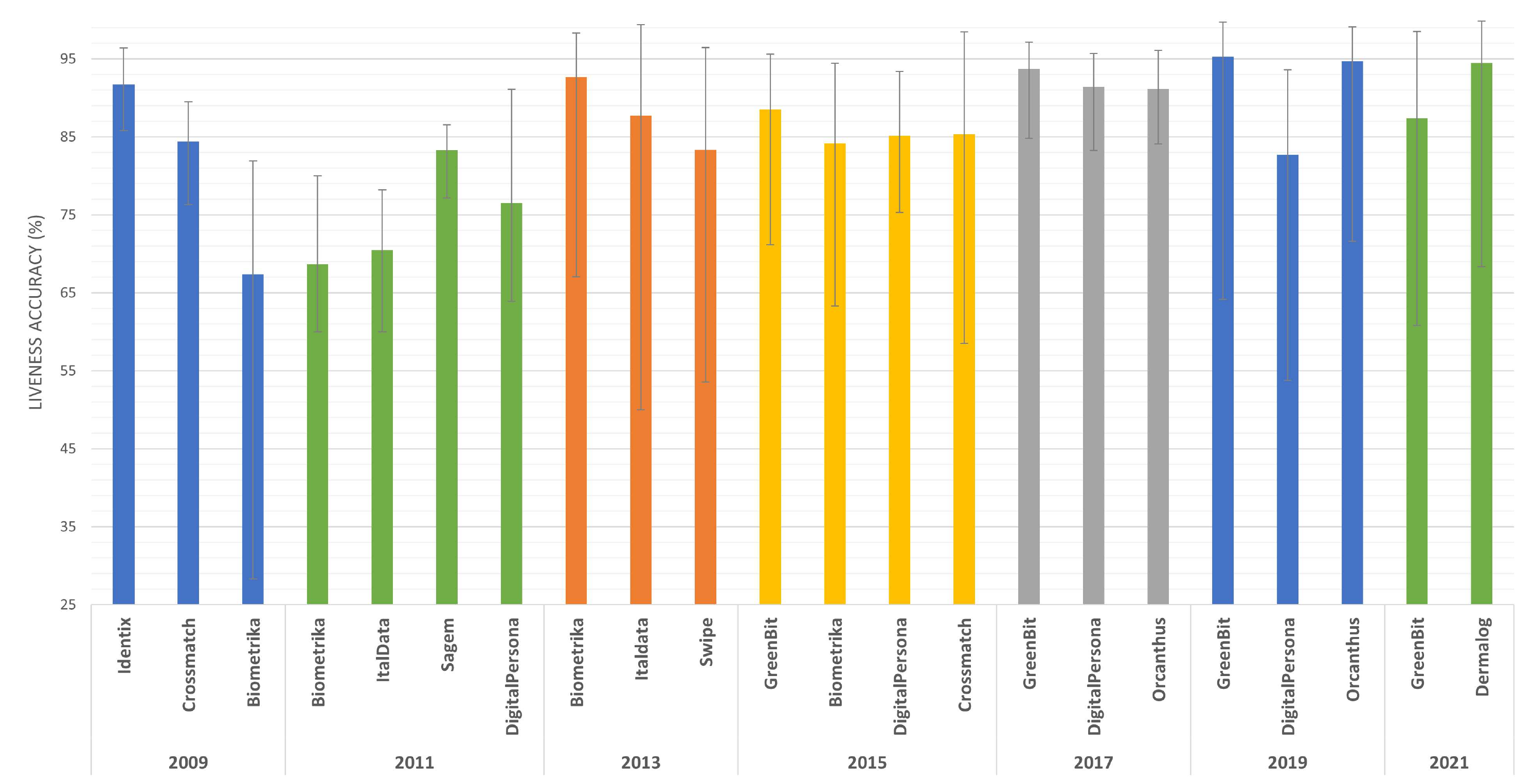}
    \caption{Liveness accuracy for each competition dataset from 2009 to 2021}
    \label{fig:sensors_liveness}
\end{figure}

The Fig. \ref{fig:sensors_liveness} summarizes the average liveness accuracy sensor-by-sensor for the datasets created by consensual method of each LivDet edition. In each bin the range between maximum and minimum accuracy is shown by whiskers. 
It is easy to observe that within each edition there is a great variability between sensors, demonstrating the great influence of the technical specifications of each of them. Obviously, this graph is to be read in light of the considerations of the composition of the datasets (e.g. consensual and non-consensual) and the challenges of each edition.

\begin{figure}
    \centering
    \includegraphics[width=\textwidth]{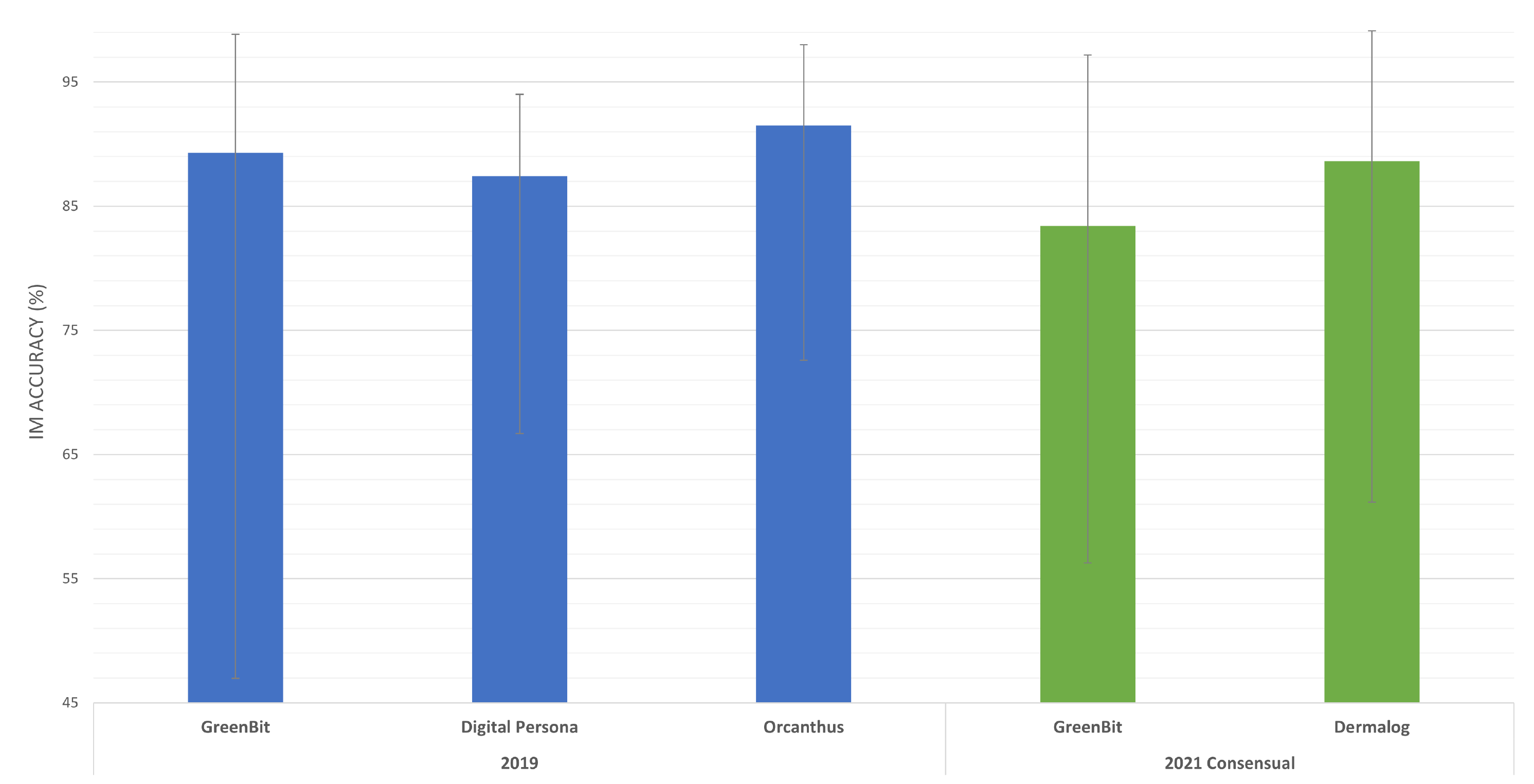}
    \caption{Integrated Matching (IM) accuracy for LivDet 2019 and LivDet 2021. Only these two editions have evaluated the performance of the integrated PAD and matcher systems.}
    \label{fig:sensors_im}
\end{figure}

Nevertheless, the joint analysis of liveness accuracy and IM accuracy and IAMPR, respectively in the Figures \ref{fig:sensors_liveness},\ref{fig:sensors_im}, highlights the following behavior: a higher IM accuracy always corresponds to a better liveness accuracy, independent of the dataset.
This suggests that the PAD system has a significant impact on the performance of the integrated system, regardless of the fusion rules and matching algorithms used.

\subsection{Non Consensual vs Consensual data}
\label{section:unconsensual}
\begin{figure}
    \centering
    \includegraphics[width=\textwidth]{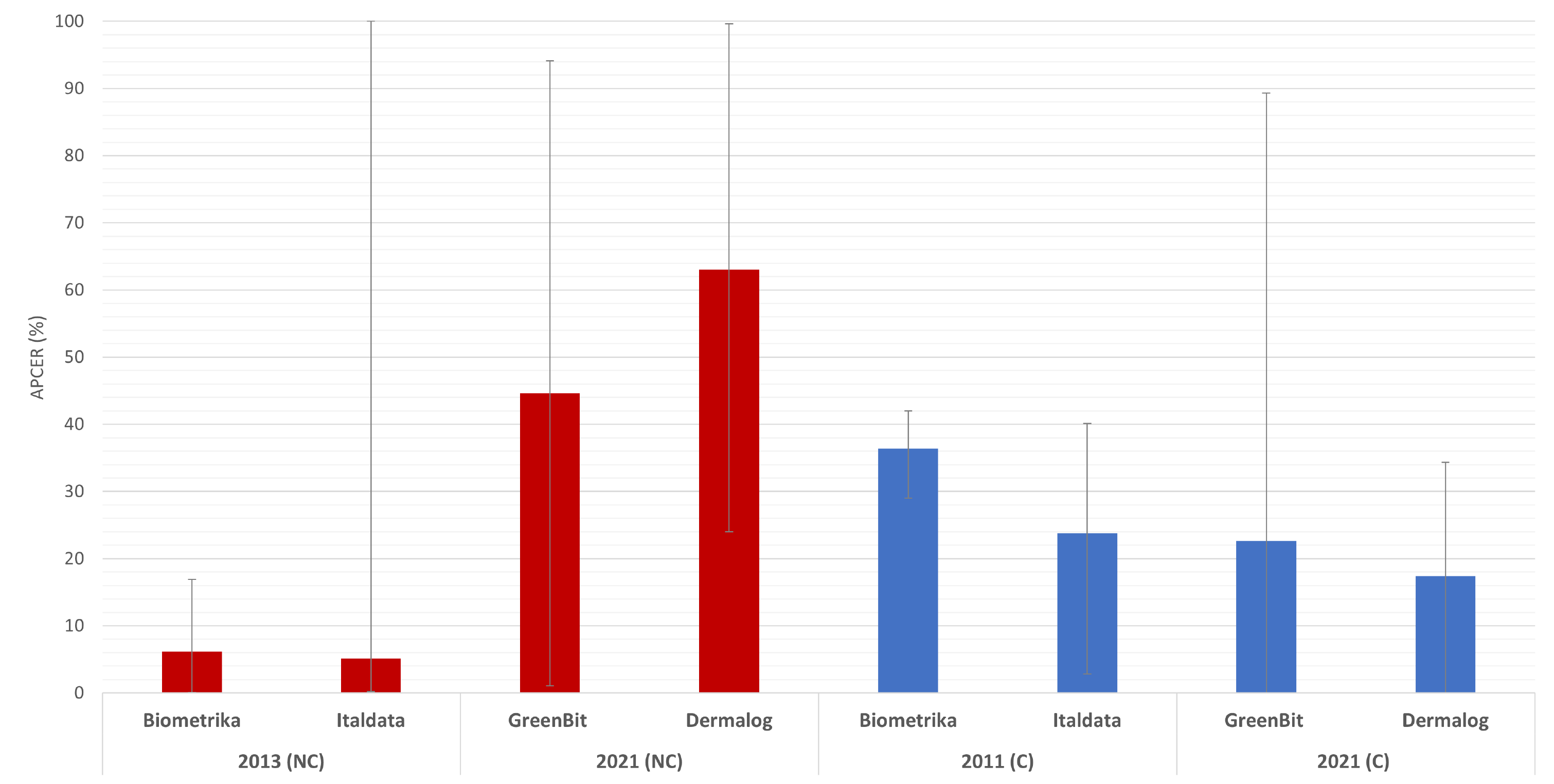}
    \caption{Comparison between APCER for consensual datasets of LivDet 2011 and 2021 (Biometrika, Italdata, GreenBit, and Dermalog) and the relative non-consensual counterpart of 2013 and 2021 editions. Minimum and maximum APCER for each sensors are reportend through whiskers.}
    \label{fig:NCvsSS}
\end{figure}

In the 2013 and 2021 editions, in addition to the consensual datasets, two non-consensual datasets were created, using latent fingerprints. The LivDet 2011 subset contains latents obtained through magnetic powder, while LivDet 2021 uses the novel ScreenSpoof technique. To verify the effectiveness of the two non-consensual (NC) methods used in the two competitions, we compared the results obtained in terms of APCER with the corresponding consensual (C) datasets (Fig. \ref{fig:NCvsSS}). As a correspondence to the non-consensual data of LivDet 2013, we chose the samples of LivDet 2011 obtained from the same sensors, Biometrika and Italdata.

Although algorithms and participants were different, the comparison between LivDet 2011 and LivDet 2013 points out that latent fingerprints acquired through the traditional non-consensual method are far from being a real threat for fingerprint PADs.
On the contrary, the attack with the ScreenSpoof method proved to be very dangerous. The high APCER is also due to the fact that the type of replication of the fingerprint was completely unknown to the PADs who had been trained only with fake obtained by the consensual method. 
The higher error rate, the lower effort to manufacture the fingerprint artefacts, and the complete unawareness of the user make this attack much more realistic than the classic consensual method and highlight the need to make future PADs ``aware'' of this eventuality.

\subsection{Materials analysis}
\label{section:materials}
\begin{figure}
    \centering
    \includegraphics[width=\textwidth]{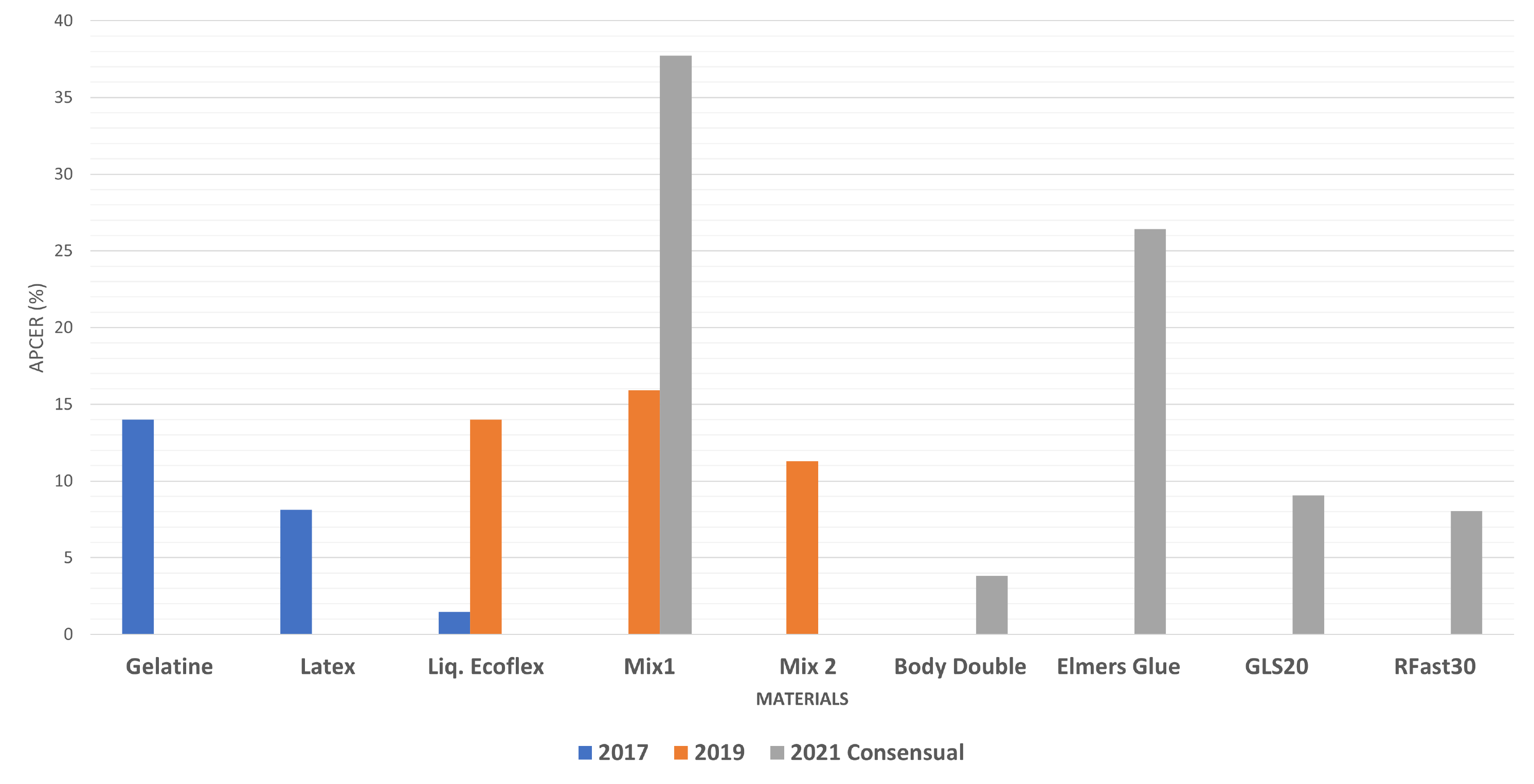}
    \caption{Comparison of APCER for test set materials from the last three editions of LivDet (consensual datasets only).}
    \label{fig:apcermaterials}
\end{figure}

The type of material used to make the spoof highly influences the success of an attack on a PAD \cite{9079541}.
To evaluate the generalization capacity of the PADs and simulate a realistic and dangerous eventuality, the test sets of the last three editions of LivDet contained only materials ``never seen before'', therefore different from those of the training sets.
Figure \ref{fig:apcermaterials} analyzes the rates of misclassified fake fingerprints for each material used in the consensual test datasets of the last three LivDet editions. 

The most immediate comparison can be made between the test materials of LivDet 2017 and LivDet 2019 that share the training set.
As always, a certain variability due to the different PADs submitted must be taken into account.

However, the behavior of the liquid ecoflex deserves attention as it turns out to be very easy to recognize in the 2017 edition while it is on average with the other materials in 2019. It is worth noting that in 2017 two characteristics helped the PADs in the classification. The first is the presence of users in the test shared with the training set to analyze the so-called user-specific effect. The second is that an inexperienced operator built part of the 2017 test set to evaluate the difference between the attackers' skills.
We, therefore, believe that these two details and the difference in PADs lead to the marked difference in APCER of the two editions for this material.

On the other hand, they do not explain the significant performance disparity for the other material in common across two editions. Mix 1 is present in the test set of LivDet 2019 and LivDet 2021. These editions present a very different training set. In particular, the training set of LivDet2021 contains only silicone and rubbers materials (Fig. \ref{fig:materials}). PADs trained with these data compositions do not learn the characteristics of the gluey materials and cannot distinguish them from live samples. In support of this thesis, we can evidence how the second most challenging material to classify is Elmer's glue. In contrast, the other three silicone materials have a lower error rate.

\begin{figure}
    \centering
    \includegraphics[width=\textwidth]{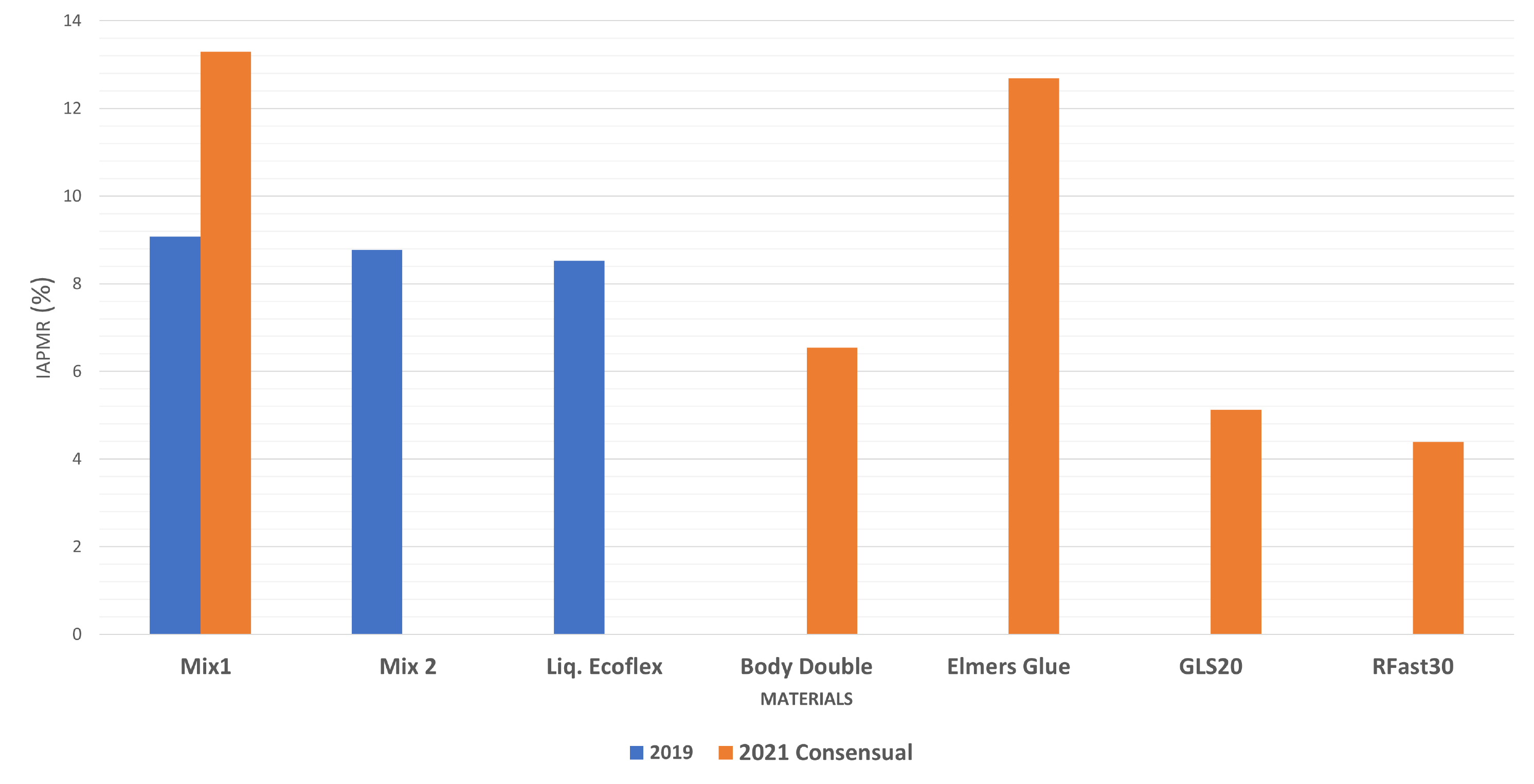}
    \caption{Comparison of IAPMRs for test set materials from the last three editions of LivDet (consensual datasets only).}
    \label{fig:materials_iapmr}
\end{figure}

To understand if the fakes made up of these two materials are actually a risk for the security of the fingerprint recognition system, we also report the results on the integrated system. Figure \ref{fig:materials_iapmr} highlights how the spoofs made with the two glue-like materials are dangerous not only from the liveness detection point of view but also manage to cheat the matcher and therefore report a high IAPMR.
Another evidence that emerges from the materials analysis is related to the Body Double: although the PADs easily recognize it, it achieves a high IAMPR. This demonstrates the high quality of spoofs that keep information relating to the individual (minutiae).



\subsection{LivDet Systems Results}

The number of submitted systems remained low for the LivDet Systems competition until the 2017 competition which saw a drastic rise in the amount of competitors (Table \ref{table:systems}). Starting from the two presented in LivDet 2011, in 2017, there were a total of 14 submitted system to the two sub-competition: nine systems submitted (with a 10th unofficial submission) for the PAD only competition and five for the integrated PAD and matcher systems competition \cite{8698578}.

\begin{table}[]
\caption{Number of systems submitted to the LivDet Systems competition.}
\label{table:systems}
\begin{tabular}{|l|c|c|c|c|c|c|c|}
\hline
\multicolumn{1}{|c|}{\textbf{LivDet Systems Edition}} & \textbf{2011} & \textbf{2013} & \textbf{2015} & \textbf{2017 PAD Only} & \textbf{2017 Integrated systems}  \\ \hline
\textbf{\# Participants}                      & 2             & 2             & 1             & 9(10)            & 5       \\ \hline    
\end{tabular}
\end{table}

Across these competitions it can be seen that the general trend of error rates has decreased with one exception. The winning system in Livdet 2013 had the best overall results from all of the Livdet Systems competitions. 

In 2011, Dermalog performed at a BPCER of 42.5\% and a APCER of 0.8\%. GreenBit performed at a BPCER of 38.8\% and a APCER of 39.47\%. Both systems had high BPCER scores. The 2013 edition produced much better results since Dermalog performed at a BPCER of 11.8\% and a APCER of 0.6\%. Anonymous1 performed at a BPCER of 1.4\% and a APCER of 0.0\%. Both systems had low APCER rates. Anonymous1 received a perfect score of 0.0\% error, successfully determining every spoof finger presented as a spoof. Anonymous2, in 2015, scored a BPCER of 14.95\% and a APCER of 6.29\% at the (given) threshold of 50 showing an improvement over the general results seen in LivDet 2011, however the anonymous system did not perform as well as what was seen in LivDet 2013. For LivDet Systems 2017 (Table \ref{table:PADres}), from the nine submissions for the PAD-only competition, Dermalog ZF1 performed the best with an APCER of 1.62\% and a BPCER of 1.33\%. The second place system is the Dermalog ZF1+ with an APCER of 2.61\% and a BPCER of 0.91\% and the third place system is the Dermalog LF10 with an APCER of 3.23\% and a BPCER of 0.82\%.  In both Livdet Systems 2015 and Livdet 2017, the most common problem with submitted systems was play-doh. In LivDet 2015, the play-doh colour contributed significantly to the system's issues with spoofs, resulting in a 28\% APCER and thus accounting for a substantial portion of the total 6.29\% APCER. In LivDet 2017, even though the Dermalog ZF1 had an APCER of 1.62 percent, it had a 10\% error rate for play-doh spoofs.

\begin{table*}[h]
\begin{center}
\caption{Error rates by competitor for LivDet Systems 2017 PAD only competition.}
\label{table:PADres}
\begin{tabular}{| l| c| c| c| }
  \hline
  System & APCER & BPCER  & ACE \\    \hline 		
F1 & 2.06 & 4.45 & 3.25 \\
ZF1 & 1.62 & 1.33 & 1.475  \\
ZF1+ & 2.61 & 0.91 & 1.76 \\
LF10 & 3.23 & 0.82 & 2.025 \\
GB & 10.08 & 1.89 & 5.985 \\
Anon1 & 7.62 & 22.6 & 15.11\\
Anon2 & 10.63 & 10.19 & 10.41 \\
Anon3 & 8.17 & 20.18 & 14.175 \\
Anon4 & 0.4 & 18.4 & 9.4 \\
  \hline  
\end{tabular}
\end{center}
\end{table*}

Overall, LivDet Systems 2017 showed a tremendous improvement in the error rates recorded from the systems. Increased in the strength of algorithms for detecting PAIs has improved throughout constant testing of systems and there is the expectation that a future LivDet Systems competition will show even more growth.

\section{Conclusion}

The LivDet competition series is becoming increasingly crucial in bringing together the most recent solutions and assessing the state of the art in fingerprint PAD. The continuous increase of competitors over the years clearly indicates the growing interest in this research topic.
This chapter presented the design and results of the seven Liveness Detection International Competitions on software-based fingerprint liveness detection methods and fingerprint systems with artifact detection capabilities. Each competition has presented new challenges to the research community, resulting in unique solutions and new insights, proving to be essential in taking fingerprint liveness detection to the next level. 
However, this heterogeneity has made the comparison of the results more challenging.
Overall, the liveness detection algorithms and systems have improved significantly over time, even under harsh conditions such as including never-before-seen materials in the test set or introducing the new ScreenSpoof fabrication method. 
On the other hand, the investigation of integrated systems performed in the last two editions revealed that PAD algorithms achieved an average accuracy of around 90\%. This implies that a typical PAD would miss 10\% of both presentation attacks and legitimate presentations. Obviously, the acceptability of this error rate depends on the application, but in general, PAD technology in this context does not appear to be fully mature.
Furthermore, the reported spoof material analysis confirmed two points: (1) the operator's skill to create and exploit counterfeit replicas has a significant impact on performance, as seen by a comparison of the 2017 and 2019 test sets; (2) since PAD systems rely mostly on the training-by-example method, the representativeness of the training set has a serious influence on their accuracy. This was especially evident in the 2021 edition, where the absence of gluey material in the training set resulted in an unexpected increase in the error rate for such materials.
In conclusion, we believe that the findings presented in this chapter will be beneficial to academics and companies working on presentation attack detection, encouraging them to improve existing solutions and uncover new ones. For our part, we will continue on this path and keep posing always new challenges to help fight this still, unfortunately, open problem.

\bibliographystyle{abbrv}
\bibliography{references}

\end{document}